\documentclass[final,5p,times,twocolumn]{elsarticle}
\usepackage{lscape}
\usepackage{amsmath}
\usepackage[colorlinks=true]{hyperref}
\usepackage{boldline}
\usepackage{capt-of}
\usepackage{csquotes}
\usepackage{placeins}
\usepackage{dirtytalk}
\usepackage{graphicx}
\usepackage{subfig}
\usepackage{multirow}
\usepackage{lipsum}
\usepackage{booktabs}
\usepackage{multicol}
\usepackage{longtable}

\usepackage{amsfonts}
\usepackage{makecell}
\usepackage{adjustbox}
\usepackage{tabularx}
\usepackage[vlines]{tabularht}
\usepackage{pbox}
\usepackage{pdflscape}
\usepackage{makecell}
\usepackage{mathtools, cuted}
\usepackage[compatibility=false]{caption}
\usepackage{pifont}
\usepackage{float}
\newcommand{\xmark}{\ding{55}}%
\usepackage{subcaption} 

\usepackage{xcolor}
\definecolor{ForestGreen}{rgb}{0.13, 0.55, 0.13}  

\newcommand*\colourcheck[1]{%
  \expandafter\newcommand\csname #1check\endcsname{\textcolor{#1}{\ding{52}}}%
}
\colourcheck{blue}
\colourcheck{green}
\colourcheck{red}

\usepackage{nomencl}
\makenomenclature

\usepackage{float}
\floatstyle{plaintop}
\restylefloat{table}

\makeatletter
\newcommand{\thickhline}{%
    \noalign {\ifnum 0=`}\fi \hrule height 1pt
    \futurelet \reserved@a \@xhline
}
\newcolumntype{"}{@{\vrule width 1pt}}
\makeatother

\newlength{\Oldarrayrulewidth}

\usepackage[linesnumbered,ruled,vlined]{algorithm2e}
\SetKwInput{KwInput}{Input}                
\SetKwInput{KwOutput}{Output}              

\usepackage[table,xcdraw]{xcolor}

\newsavebox\MBox

\usepackage{amsmath}
\usepackage{amsmath}

\usepackage{tikz}

\biboptions{sort&compress}

\journal{Medical Image Analysis}

\begin{document}

\begin{frontmatter}

\title{Prompt-Conditioned Channel Attention for Hierarchical Feature Modulation toward Anatomy-Agnostic Segmentation}

\author[label1]{Mosharof Hossain}
\ead{mosharof28ndc@gmail.com}

\author[label2]{Md Rabiul Islam}
\ead{rabiul\_islam@tamu.edu}

\author[label1]{Limon Halder}
\ead{limonhalder79@gmail.com}

\author[label2]{Erchin Serpedin}
\ead{eserpedin@tamu.edu}

\author[label1,label3]{Md Kamrul Hasan\corref{cor1}\fnref{fn1}}
\ead{k.hasan22@imperial.ac.uk OR kamruleeekuet@gmail.com}

\address[label1]{Department of Electrical and Electronic Engineering, Khulna University of Engineering \& Technology (KUET), Khulna-9203, Bangladesh}
\address[label2]{Department of Electrical and Computer Engineering, Texas A\&M University, College Station, TX, USA}
\address[label3]{Department of Bioengineering, Imperial College London, London SW7 2AZ, UK}

\cortext[cor1]{Corresponding author}
\fntext[fn1]{Senior author.}

\begin{abstract}
\textnormal{Accurate and anatomically plausible medical image segmentation remains challenging due to low contrast, ambiguous boundaries, inter-patient variability, and modality-specific artifacts. Interactive segmentation using spatial prompts has emerged as a promising strategy to guide attention and improve localization, particularly in low-contrast or structurally ambiguous regions. However, existing methods typically restrict prompt integration to late-stage fusion and lack explicit mechanisms for channel-wise feature modulation, limiting their ability to capture deeper contextual and modality-specific variations across hierarchical representations. To address these limitations, we introduce \emph{Prompt-Conditioned Channel Attention} (PCCA), a novel modulation mechanism that enables deep, hierarchical integration of semantic prompts within encoder-decoder segmentation networks. PCCA extracts compact channel descriptors from both image and prompt features via pooling, projects them into a shared latent space, and fuses them through a gated excitation unit to compute prompt-aware channel attention weights. These weights perform hierarchical channel-wise attention by adaptively reweighting feature responses at every network stage, enabling spatially selective and semantically enriched representations. Building on this mechanism, we propose \textbf{PROMISE-Net} (\textbf{PRO}mpt \textbf{M}odulated \textbf{I}ntegration for \textbf{SE}gmentation), instantiated in two architectural variants: a convolutional model (\textbf{PROMISE-C\textsubscript{NN}}) and a transformer-based model (\textbf{PROMISE-T\textsubscript{xformer}}). Across ISIC-Lesion, Kvasir-Polyp, CAMUS-Cardiac, and Kvasir-Instrument benchmarks, integrating PCCA into PROMISE-C\textsubscript{NN} yielded relative IoU improvements of \textit{10.4}\%, \textit{8.7}\%, \textit{0.8}\%, and \textit{3.4}\% over baseline U-Net, while PROMISE-T\textsubscript{xformer} achieved corresponding gains of \textit{7.6}\%, \textit{23}\%, \textit{2.1}\%, and \textit{1.1}\% over baseline UNETR. These results demonstrate consistent cross-architectural, cross-modal, and cross-anatomical improvements, establishing PCCA coupled PROMISE-Net as a scalable and generalizable framework for prompt-aware feature modulation in medical image segmentation. The code is available at \url{https://github.com/kamruleee51/PROMISENet}.}
\end{abstract}

\end{frontmatter}

\section{Introduction}
\label{sec:introduction}
Medical image segmentation plays a crucial role in modern healthcare, offering significant benefits in clinical diagnosis, surgical planning, and treatment monitoring \cite{LITJENS201760}. Accurately delineating anatomical structures and pathological regions enables more informed clinical decision-making and contributes to improved patient outcomes. Nevertheless, achieving reliable segmentation remains highly challenging due to the inherent complexity and variability of medical images. These challenges arise primarily from two major aspects. First, medical images exhibit considerable diversity in the shape, size, and spatial configuration of anatomical structures. For instance, in skin lesion analysis, imaging artifacts such as hair, blood vessels, and uneven illumination can obscure lesion boundaries and confound interpretation. In addition, irregular contours and heterogeneous intensity distributions further degrade the performance of conventional shape- or intensity-based models \cite{8620285, gutman2016skin}. Second, image resolution exerts a notable influence on segmentation performance. Low-resolution scans often suffer from poor boundary definition and loss of fine anatomical detail, whereas excessively high-resolution images may amplify noise and introduce unstable feature responses.

\begin{figure*}[!t]
\centering
\includegraphics[width=0.9\textwidth]{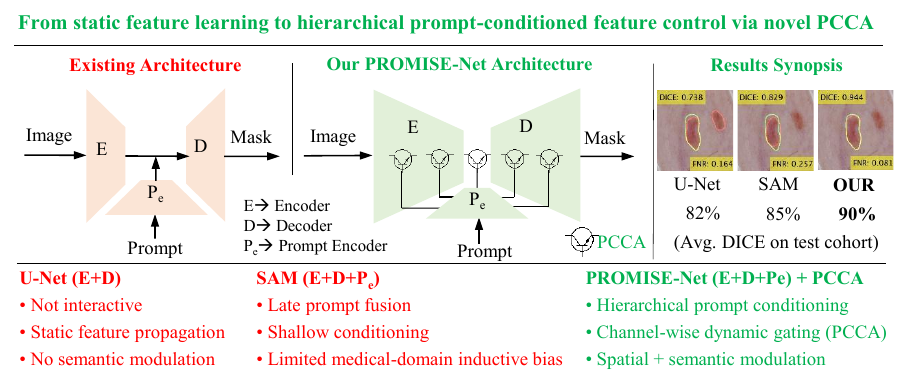}
\caption{From static feature learning to hierarchical prompt-conditioned feature control. Compared with U-Net and SAM, which rely on static representations or shallow prompt fusion, PROMISE-Net integrates Prompt-Conditioned Channel Attention (PCCA) throughout the network, enabling dynamic channel-wise gating and spatial–semantic modulation for improved segmentation performance.}
\label{fig:Graphical_Abstract}
\end{figure*}

Manual delineation by clinical experts, though highly accurate, is both labor-intensive and time-consuming, rendering it impractical for large-scale clinical deployment. To mitigate this, a wide range of automatic segmentation algorithms have been developed over the past decades \cite{hasan2021dermo,hasan2025motion}. Early approaches relied on classical image processing techniques such as thresholding \cite{wang2015hybrid} and edge detection \cite{patil2013medical}. However, these methods lacked robustness and generalizability when confronted with the high anatomical variability and complex appearance patterns present in real-world medical images. Recent advances in deep learning (DL), particularly the advent of convolutional neural networks (CNNs) \cite{ronneberger2015u} and Transformers \cite{hatamizadeh2022unetr}, have revolutionized medical image segmentation, achieving state-of-the-art results across diverse modalities and anatomical targets \cite{chen2022recent, zhang2023anatomy}. Despite their success, these models typically demand large-scale, fully annotated datasets to attain optimal accuracy and generalization. Acquiring such expert annotations is both costly and time-intensive, as it requires domain-specific knowledge and meticulous manual effort.

To alleviate the reliance on dense annotations, weakly and semi-supervised segmentation strategies have been widely explored, leveraging sparse or noisy supervisory signals such as image-level tags, bounding boxes, scribbles, or points \cite{chen2022c, kuang2023cluster, zhai2023pa, wei2023weakpolyp, zhang2022cyclemix, li2024scribformer}. Although these methods substantially reduce annotation costs, their limited spatial supervision often leads to inaccurate boundary delineation and structural inconsistencies, particularly in low-contrast or artifact-prone regions. To mitigate these shortcomings, several studies have incorporated structural priors, adversarial shape regularization, or consistency constraints to improve spatial coherence \cite{valvano2021learning, zhang2022shapepu}. Yet, these techniques primarily enforce implicit shape regularity and fail to provide explicit spatial guidance that adapts to heterogeneous imaging conditions and anatomical variability.

Among the weakly supervised paradigms, scribble supervision has proven particularly effective due to its ease of use and practicality for delineating complex anatomical structures \cite{tajbakhsh2020embracing}. Annotators need only draw a few representative strokes per class, drastically reducing annotation time. However, the sparse and localized nature of scribble annotations provides limited supervision in low-contrast or ambiguous regions, often leading to segmentation errors in internal substructures and degraded boundary precision. Conventional architectures such as U-Net struggle to reconstruct missing contextual information, resulting in incomplete shape recovery of critical anatomical components. These challenges underscore the need for frameworks capable of learning effectively from sparse supervision while preserving detailed anatomical morphology.

To strengthen the supervision signal, pseudo-labeling (PL) has emerged as an effective enhancement within scribble-guided segmentation. PL methods exploit model predictions to generate provisional labels for unlabeled regions, thereby expanding the effective training set \cite{luo2022scribble, wang2023s, lei2024pclmix}. Typically, pseudo-labels are aggregated from multiple networks or decoder branches to improve robustness. For instance,  Luo et al. \cite{luo2022scribble} dynamically blend outputs from dual decoders to diversify pseudo-label generation, while  Wu et al. \cite{wu2023compete} select high-confidence pixels from ensemble confidence maps. Similarly, Han et al. \cite{han2023scribble} employ multi-branch architectures with varied dilation rates and feature-level perturbations, and PacingPseudo \cite{yang2024non} enforces prediction consistency through a twin-network design. In addition, \cite{zhang2024scribble} integrates class activation maps to inject class-specific cues derived from scribble annotations. Despite these advances, most PL frameworks rely on simple averaging or random weighting during aggregation, which can amplify uncertainty and introduce label noise \cite{wang2021uncertainty}. As a result, pseudo-labels often lack structural coherence and fail to guide fine-grained boundary refinement effectively.

In parallel, interactive and prompt-based segmentation has gained momentum as a powerful means of injecting spatial priors directly into deep models. In this paradigm, users provide minimal yet informative cues such as points, bounding boxes, or scribbles to guide model attention and localization \cite{kirillov2023segment} (shown in Fig.~\ref{fig:Graphical_Abstract}). This strategy is particularly attractive for human-in-the-loop clinical workflows, where clinicians can iteratively refine results while preserving the efficiency of automated inference. Despite these advantages, most existing prompt-driven frameworks, including SAM \cite{kirillov2023segment}, incorporate prompts only through late-stage fusion or single-layer conditioning, confining their influence to shallow or bottleneck representations. Moreover, the mechanisms by which prompt features interact with mainstream image features, especially at the level of channel-wise modulation, are often ad hoc or underspecified. As a result, prompt semantics are not effectively propagated across the network hierarchy, limiting the model’s ability to shape intermediate representations. This restriction hinders the capture of multi-scale contextual relationships, modality-specific artifacts, and fine anatomical structures, particularly in low-contrast or structurally ambiguous regions.

In this work, we propose the following key contributions. First, we introduce \emph{Prompt-Conditioned Channel Attention (PCCA)}, a novel attention mechanism that enables deep, hierarchical modulation of spatial prompts within encoder–decoder segmentation networks, allowing semantic guidance to propagate consistently throughout the feature hierarchy (illustrated in Fig.~\ref{fig:Graphical_Abstract}). Unlike existing prompt-driven designs that condition only shallow or bottleneck layers, PCCA embeds prompts directly into the feature transformation process at every stage. PCCA is a lightweight, prompt-conditioned channel excitation mechanism that internally executes squeeze-and-excitation operations to adaptively fuse prompt-derived spatial priors with visual features, thereby enhancing region-of-interest focus, boundary localization, and anatomical consistency while preserving computational efficiency. Second, we integrate PCCA into a unified segmentation framework, \textbf{PROMISE-Net} (\textbf{PRO}mpt \textbf{M}odulated \textbf{I}ntegration for \textbf{SE}gmentation), instantiated in two complementary variants: a convolutional model (\textbf{PROMISE-C\textsubscript{NN}}) and a transformer-based model (\textbf{PROMISE-T\textsubscript{xformer}}). This design demonstrates the cross-architectural generality and plug-in flexibility of the proposed mechanism. Finally, extensive evaluations on the ISIC-Lesion, CAMUS-Cardiac, Kvasir-Polyp, and Kvasir-Instrument benchmarks show that PROMISE-Net consistently achieves cross-architectural, cross-modal, and cross-anatomical improvements over strong state-of-the-art baselines. Collectively, these contributions establish PCCA coupled PROMISE-Net as a scalable and generalizable framework for prompt-aware feature modulation in medical image segmentation.

\begin{figure*}[!ht]
\centering
\includegraphics[width=\textwidth, trim=15 10 10 0, clip]{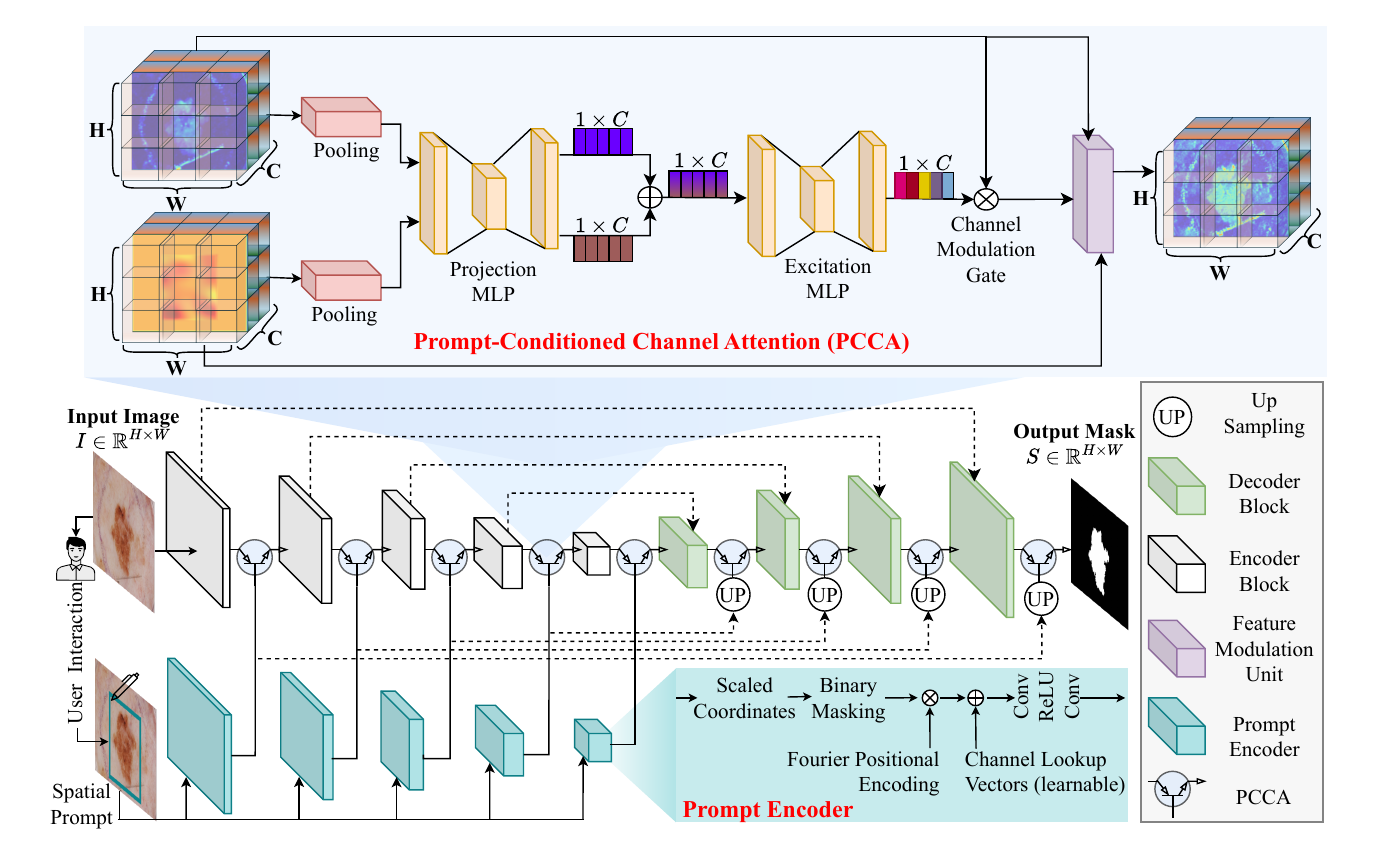}
\caption{Overview of PROMISE-Net architecture. The top panel presents the proposed PCCA module, while the bottom panel illustrates the segmentation backbone with hierarchical PCCA integration. Embedding PCCA within convolutional blocks yields PROMISE-C\textsubscript{NN}, while analogous integration in transformer blocks results in PROMISE-T\textsubscript{xformer}.}
\label{fig:network}
\end{figure*}

\section{PROMISE-Net Framework}
Building upon the motivation outlined in Section~\ref{sec:introduction}, we propose PROMISE-Net, a prompt-aware segmentation framework composed of an image encoder, a prompt encoder, and hierarchical modulation units based on the PCCA module, as illustrated in Fig.~\ref{fig:network}. At its core, PCCA enables deep, hierarchical integration of semantic prompts within encoder–decoder networks. Specifically, it operates on intermediate feature maps from the image encoder and semantic embeddings from a lightweight prompt encoder to generate spatially modulated representations, wherein PCCA acts as a gating mechanism that regulates the flow of mainstream visual features according to prompt relevance.

The gating behavior of PCCA is analogous to that of an electronic switching transistor: the encoder feature map acts as the primary input (collector current), while the prompt embedding provides the control signal (base drive). This relationship is formalized as:
\begin{equation}
\label{gate}
\mathbf{Y} = \sigma(\mathbf{P}) \odot \mathbf{F},
\end{equation}
where $\mathbf{F}$ denotes the encoder feature map, $\sigma(\mathbf{P})$ is the prompt-conditioned gate broadcast over $H{\times}W$, and $\mathbf{Y}$ is the modulated output. When prompt activation is strong, $\sigma(\mathbf{P})\approx1$ and the gate conducts; when prompt activation is weak or absent, $\sigma(\mathbf{P})\approx0$, suppressing non-salient responses. This transistor-like modulation dynamically amplifies or attenuates spatial activations, thereby focusing the network on anatomically relevant regions.

Compact channel descriptors are extracted from both visual and prompt streams via global average pooling, projected onto a shared latent space, and fused through a gated excitation unit to produce channel-wise modulation weights that adaptively reweight image features at every hierarchical level. Through this design, PCCA preserves spatial semantics while adapting to hierarchical context. PROMISE-Net is designed in two architectural variants: a convolutional model (\textbf{PROMISE-C\textsubscript{NN}}) and a transformer-based model (\textbf{PROMISE-T\textsubscript{xformer}}), demonstrating cross-architectural generality. The following subsections describe each component in detail.

\subsection{Encoder Representation Learning}
Segmentation requires hierarchical feature representations that capture both low-level texture details and high-level semantic cues. To this end, the encoder $\mathcal{E}$ forms the backbone of PROMISE-Net, progressively abstracting visual information into multi-scale latent features that serve as the basis for prompt-conditioned modulation. Given an input image $I^n \in \mathbb{R}^{H \times W}$, the network predicts a segmentation mask $S^n \in \mathbb{R}^{H \times W}$ through the learnable mapping $\mathcal{F}$: 
\begin{equation}
S^n = \mathcal{F}(I^n; \theta), \quad n = 1,\dots,N,
\label{eq:segmentation_main}
\end{equation}
where $\theta$ denotes the learnable parameters of the segmentation network ($\mathcal{F}$). The encoder extracts progressively higher-level features through convolutional or transformer operations, as shown in~\eqref{eq:encoder_features}:
\begin{equation}
\begin{aligned}
F^{0} &= I^n, \\
F^{l} &= \sigma\big(\mathrm{BN}(\mathrm{Conv}_{3\times3}(F^{l-1}))\big), \quad l = 1,\dots,L, \\
F^{l} &= \mathrm{Down}(F^{l}),
\end{aligned}
\label{eq:encoder_features}
\end{equation}
where $\mathrm{Conv}_{3\times3}$, $\mathrm{BN}$,  $\sigma(\cdot)$ and $\mathrm{Down}(\cdot)$ stand for  $3{\times}3$ convolution, batch normalization,  nonlinear activation function (ReLU), and spatial downsampling, respectively.  
As the network depth increases ($l \to L$), the feature dimensionality expands (64, 128, 256, 512, 1024), enabling the encoder to capture appearance, shape, and boundary cues across multiple spatial scales.

In \textbf{PROMISE-C\textsubscript{NN}}, the hierarchy in~\eqref{eq:encoder_features} is purely convolutional, excelling at capturing local, fine-grained details such as edges and textures.  
Conversely, \textbf{PROMISE-T\textsubscript{xformer}} replaces convolutional blocks with transformer encoders to model long-range spatial dependencies and enable global contextual reasoning \cite{hatamizadeh2022unetr}, as formalized in~\eqref{eq:transformer_block}:
\begin{equation}
F^{l} = \mathrm{MSA}\big(\mathrm{LN}(F^{l-1})\big) + \mathrm{MLP}\big(\mathrm{LN}(F^{l-1})\big),
\label{eq:transformer_block}
\end{equation}
where $\mathrm{MSA}$, $\mathrm{MLP}$,  and $\mathrm{LN}$ denote the multi-head self-attention,  feed-forward projection, and  layer normalization, respectively.  
Together, these two architectural variants enable the PCCA-coupled PROMISE-Net to evaluate the cross-architectural generality of the proposed modulation mechanism, providing a unified foundation for prompt-aware feature integration.

\subsection{Prompt Encoder}
The Prompt Encoder transforms a coarse bounding box, provided by the user, into a dense spatial prompt that aligns with the resolution and semantics of the encoder features.  
Unlike binary masks that merely indicate the region of interest, this design incorporates spatial location, positional context, and semantic embedding, enabling the model to learn both \emph{where} the target region lies and \emph{how} it interacts with the hierarchical feature representations of the backbone.

Given an input bounding box $(x_1, y_1, x_2, y_2)$ defined in image coordinates with spatial dimensions $(H_i, W_i)$, the coordinates are normalized to match the spatial resolution of the corresponding feature map $(H_f, W_f)$  as follows: 
\begin{equation}
(x_1', y_1', x_2', y_2') = 
\Big(
\tfrac{W_f}{W_i}x_1,\;
\tfrac{H_f}{H_i}y_1,\;
\tfrac{W_f}{W_i}x_2,\;
\tfrac{H_f}{H_i}y_2
\Big),
\label{eq:coordnorm}
\end{equation}
ensuring that the prompt aligns precisely with the spatial scale of the intermediate feature representation.  
A binary mask $\mathbf{M}_b \in \{0,1\}^{H_f\times W_f}$ is then generated to localize the region of interest: 
\begin{equation}
\mathbf{M}_b(i,j) =
\begin{cases}
1, & x_1' \le j < x_2',\;\; y_1' \le i < y_2',\\
0, & \text{otherwise,}
\end{cases}
\label{eq:boxmask}
\end{equation}

To enrich spatial priors beyond binary localization, we apply a random Fourier positional encoding that maps normalized spatial coordinates $\mathbf{U}(i,j) = [\,j/W_f,\, i/H_f\,]$ into a high-dimensional sinusoidal space, as defined in~\eqref{eq:fourier} and inspired by prior work~\cite{kirillov2023segment,tancik2020fourier}:
\begin{equation}
\label{eq:fourier}
\begin{aligned}
\mathbf{Z}(i,j)
&= 2\pi\,\mathbf{U}(i,j)\mathbf{G}, \\
\mathbf{E}(i,j)
&= \left[
\sin\!\big(\mathbf{Z}(i,j)\big),
\;
\cos\!\big(\mathbf{Z}(i,j)\big)
\right].
\end{aligned}
\end{equation}
where $\mathbf{G}\in\mathbb{R}^{2\times D}$ is a fixed Gaussian projection matrix.  
The resulting encoding $\mathbf{E}\in\mathbb{R}^{H_f\times W_f\times 2D}$ captures smooth positional variations and introduces global spatial awareness, analogous to positional encodings used in vision transformers \cite{dosovitskiy2020image}. 
Each bounding box is further assigned a learnable semantic identity through an embedding vector $\mathbf{e}_{\text{box}}\in\mathbb{R}^{C}$.  
This embedding is broadcast spatially and fused with the masked positional encoding to produce the preliminary dense prompt representation:  
\begin{equation}
\mathbf{F}_b = \big(\mathbf{E}\odot\mathbf{M}_b\big) + \mathbf{e}_{\text{box}}\otimes\mathbf{1}_{H_f\times W_f},
\label{eq:fbox}
\end{equation}
where $\odot$ denotes elementwise multiplication and $\otimes$ represents outer broadcasting.  
Equation~\eqref{eq:fbox} ensures that the encoded prompt emphasizes the target region while introducing a learnable semantic bias that distinguishes bounding-box prompts from other spatial tokens.
To enforce local smoothness and reduce box-edge discontinuities, the preliminary prompt representation $\mathbf{F}_b$ is refined through a two-layer convolutional block with shared weights across channels as follows: 
\begin{equation}
\hat{\mathbf{F}}_b = 
\mathrm{Conv}_{3\times3}
\Big(
\mathrm{ReLU}\big(
\mathrm{Conv}_{3\times3}(\mathbf{F}_b)
\big)
\Big),
\label{eq:convref}
\end{equation}
producing a dense and continuous feature map $\hat{\mathbf{F}}_b \in \mathbb{R}^{C\times H_f\times W_f}$.  
This refinement acts as a spatial low-pass filter, promoting continuity across bounding-box boundaries and suppressing high-frequency artifacts. For each level $l$ of the encoder-decoder hierarchy with spatial resolution $(H_l, W_l)$ and channel dimension $C_l$, the refined prompt is projected to a matching scale: 
\begin{equation}
\mathbf{P}^l = 
\mathrm{Resize}\big(\hat{\mathbf{F}}_b\big)
\in
\mathbb{R}^{C_l\times H_l\times W_l},
\label{eq:promptscale}
\end{equation}
where $\mathrm{Resize}(\cdot)$ denotes bilinear interpolation followed by a $1{\times}1$ projection (\enquote{\textbf{UP}} block in Fig.~\ref{fig:network}).  
These multi-scale prompt features $\{\mathbf{P}^l\}_{l=1}^{L}$ are subsequently injected into both the encoder and decoder through prompt-conditioned channel attention (Section~\ref{pcca}).

\subsection{Prompt-Conditioned Channel Attention (PCCA)}
\label{pcca}
To integrate spatially embedded priors derived from bounding-box prompts, as described in the previous section, with image features in a semantically consistent and scale-adaptive manner, we introduce the \emph{PCCA} module. Unlike naïve concatenation or additive fusion, PCCA modulates channel activations through prompt-aware excitation, enabling the network to dynamically emphasize channels that are most informative for the user-specified region of interest.

Let the image feature map at stage $l$ be denoted as $\mathbf{F}_b \in \mathbb{R}^{B \times C \times H \times W}$, and let the corresponding prompt embedding be $\mathbf{F}_p \in \mathbb{R}^{B \times D \times H \times W}$.  
To summarize global contextual information from both streams, pooled descriptors are extracted via global average pooling: 

\begin{equation}
\label{eq:avgpool}
\begin{aligned}
\mathbf{v}_b
&=
\frac{1}{HW}
\sum_{i=1}^{H}
\sum_{j=1}^{W}
\mathbf{F}_b[:,:,i,j], \\
\mathbf{v}_p
&=
\frac{1}{HW}
\sum_{i=1}^{H}
\sum_{j=1}^{W}
\mathbf{F}_p[:,:,i,j].
\end{aligned}
\end{equation}

Both pooled descriptors are projected into a shared latent space of dimension $C$ using a projection MLP: 
\begin{equation}
\tilde{\mathbf{v}}_b = W_b\,\mathbf{v}_b, 
\qquad
\tilde{\mathbf{v}}_p = W_p\,\mathbf{v}_p,
\label{eq:proj}
\end{equation}
where $W_b\in\mathbb{R}^{C\times C}$ and $W_p\in\mathbb{R}^{D\times C}$ are learnable linear transformations that align visual and prompt semantics within a common latent representation.  
A joint modulation descriptor is then obtained via additive fusion: 
\begin{equation}
\mathbf{z} = \tilde{\mathbf{v}}_b + \tilde{\mathbf{v}}_p,
\label{eq:fuse}
\end{equation}
which aggregates complementary information from image and prompt cues.  
The combined descriptor $\mathbf{z}$ is subsequently passed through a lightweight bottleneck gating function implemented as an excitation MLP (inspired by \cite{hu2018squeeze}): 
\begin{equation}
\mathbf{s} = \sigma\big(W_2\,\delta(W_1\,\mathbf{z})\big).
\label{eq:mlp}
\end{equation}
where $W_1\in\mathbb{R}^{C\times C/r}$ and $W_2\in\mathbb{R}^{C/r\times C}$ form a two-layer MLP with reduction ratio $r$, $\delta(\cdot)$ denotes the ReLU activation, and $\sigma(\cdot)$ is the sigmoid gating function.

Equation~\eqref{eq:mlp} yields a channel-wise excitation vector $\mathbf{s}\in[0,1]^C$ that adaptively reweights image feature channels according to prompt-conditioned relevance.  
The modulated feature map is  reconstructed as:  
\begin{equation}
\mathbf{F}_{\text{out}} 
= \mathbf{F}_b \odot \mathbf{s} + \mathbf{F}_b \odot \mathbf{F}_p,
\label{eq:output}
\end{equation}
where $\odot$ denotes elementwise multiplication with broadcasting. Equation \eqref{eq:output} encapsulates the \enquote{\textbf{Feature Modulation Unit}} in Fig.~\ref{fig:network}, while  $\mathbf{F}_b \odot \mathbf{s}$ defines the \enquote{\textbf{Channel Modulation Gate}}. The channel-wise excitation term provides global prompt-conditioned feature selection, while the pixel-wise interaction term complements it by enforcing spatially localized modulation, ensuring that prompt guidance remains both semantically selective and spatially precise.

At inference, the excitation vector $\mathbf{s}$ (in~\eqref{eq:mlp}) behaves analogously to the control current of a transistor: when a strong prompt activation $\mathbf{v}_p$ (in~\eqref{eq:avgpool}) is present, $\mathbf{s}$ opens the gate to amplify feature flow; when prompt activation is weak or absent, $\mathbf{s}$ suppresses non-relevant channels, preventing unwanted activations in unrelated regions.  
Accordingly, the overall gating behavior of PCCA is compactly expressed as: 
\begin{equation}
\label{eq:pcca_final}
\begin{aligned}
\mathbf{F}_{\text{out}}
&= \mathbf{F}_b \odot g(\mathbf{v}_p), \\
g(\mathbf{v}_p)
&= \sigma\!\left(
W_2\,\delta\!\left(
W_1\left(
W_b\mathbf{v}_b + W_p\mathbf{v}_p
\right)
\right)
\right)
+ \mathbf{F}_p.
\end{aligned}
\end{equation}
where $g(\mathbf{v}_p)$ denotes a prompt-conditioned gating function that combines channel-wise excitation with spatially localized modulation.  
This generalizes classical squeeze-and-excitation~\cite{hu2018squeeze} by conditioning the excitation function on both visual and prompt embeddings rather than visual statistics alone.

PCCA modules are embedded at all encoder and decoder levels, ensuring that prompt guidance percolates throughout the network hierarchy. This hierarchical conditioning enables spatially adaptive, semantically consistent, and anatomically plausible feature modulation, allowing the model to maintain focus on clinically relevant structures across scales.

\subsection{Prompt-Conditioned Fusion and Decoder Reconstruction}
The decoder $\mathcal{D}$ complements the encoder by hierarchically reconstructing spatial details while preserving prompt-guided semantic focus. At each decoder stage, upsampled feature maps are fused with their corresponding encoder features, both already modulated by PCCA, thereby ensuring bidirectional propagation of prompt information throughout the hierarchy.

At each encoder stage $l$, prompt integration is performed via prompt-conditioned channel attention: 
\begin{equation}
\hat{\mathbf{F}}^{l} = \mathrm{PCCA}\big(\mathbf{F}^{l}, \mathbf{P}^l\big),
\label{eq:encoder_pcca}
\end{equation}
where $\hat{\mathbf{F}}^{l}$ denotes the prompt-modulated encoder feature at level $l$.  
During decoding, we denote by $\mathbf{G}^{l}$ the feature map reconstructed at the $l^{\text{th}}$ decoder stage.  
Prompt information continues to guide feature refinement through successive fusion and modulation operations: 
\begin{equation}
\label{eq:decoder_update_pcca}
\begin{aligned}
\mathbf{G}^{L} &= \hat{\mathbf{F}}^{L}, \\
\tilde{\mathbf{G}}^{l-1}
&= \psi\left(
\mathrm{Up}\left(\mathbf{G}^{l}\right)
\Vert
\hat{\mathbf{F}}^{l-1}
\right), \\
\hat{\mathbf{G}}^{l-1}
&=
\mathrm{PCCA}\left(
\tilde{\mathbf{G}}^{l-1},
\mathbf{P}^{l-1}
\right),
\qquad l = L,\dots,1.
\end{aligned}
\end{equation}
\noindent where $\mathrm{Up}(\cdot)$ denotes bilinear upsampling or transposed convolution, and $\psi(\cdot)$ represents a $3{\times}3$ convolution followed by normalization and nonlinear activation.  

This dual-stage prompt integration enables the encoder and decoder to co-adapt visual evidence and spatial intent, leading to region-specific reconstruction and reduced leakage into irrelevant anatomical areas.  
The final segmentation mask is obtained: 
\begin{equation}
S = \sigma\big(\mathrm{Conv}_{1\times1}(\hat{\mathbf{G}}^{0})\big) ,
\label{eq:segmentation_output}
\end{equation}
yielding a dense probability map whose boundaries closely align with the user-indicated regions.

\subsection{Unified Forward Formulation and Implementation}
Combining prompt-conditioned encoding and reconstruction, the complete PROMISE-Net forward formulation is expressed as: 

\begin{equation}
\label{eq:overall_forward}
\begin{aligned}
S
&= \sigma\!\left(
\mathcal{D}_{\mathrm{PCCA}}\!\left(
\{\hat{\mathbf{F}}^{l}\},
\{\mathbf{P}^{l}\};
\theta_{\mathcal{D}}
\right)
\right), \\
\hat{\mathbf{F}}^{l}
&= \mathrm{PCCA}\!\left(
\mathcal{E}^{l}\!\left(
I; \theta_{\mathcal{E}}
\right),
\mathbf{P}^{l};
\theta_{\mathrm{PCCA}}
\right).
\end{aligned}
\end{equation}
where prompt-conditioned feature modulation is applied at every encoder stage and propagated through the decoder hierarchy.
This formulation captures the bidirectional propagation of spatial priors throughout the encoder-decoder architecture, ensuring that anatomical cues influence both feature abstraction and pixel-level reconstruction.  
Such dual-stage conditioning enables PROMISE-Net to remain interpretable, responsive to user guidance, and robust across anatomical structures, backbone architectures, and imaging modalities.

From an interpretive standpoint, each PCCA unit can be viewed as a soft gating mechanism that dynamically regulates feature conduction according to prompt activation: 
\begin{equation}
\label{eq:transistor_form}
\begin{aligned}
\hat{\mathbf{F}}^{l}
&= g\!\left(\mathbf{P}^{l}\right)
\odot \mathbf{F}^{l}, \\
g\!\left(\mathbf{P}^{l}\right)
&= \sigma\!\left(
\mathbf{W}_{2}^{l}\,
\delta\!\left(
\mathbf{W}_{1}^{l}\,
\mathrm{GAP}\!\left(\mathbf{P}^{l}\right)
\right)
\right).
\end{aligned}
\end{equation}
where $g(\mathbf{P}^l)$ denotes a prompt-dependent conduction gate.  
When the prompt strongly activates relevant spatial regions, $g(\mathbf{P}^l)\approx1$ allows unhindered information flow; conversely, when the prompt is weak or absent, $g(\mathbf{P}^l)\approx0$ suppresses non-salient activations.  
This transistor-like gating analogy provides an intuitive physical interpretation of PCCA, illustrating how prompt-conditioned modulation adaptively amplifies or attenuates neural feature dynamics across the network hierarchy.
The implementation of PROMISE-Net is summarized in Algorithm~\ref{alg:promisenet_pcca} using PyTorch-style pseudocode.

\begin{algorithm}[!t]
\caption{PROMISE-Net's PyTorch-style pseudocode.}
\label{alg:promisenet_pcca}
\textcolor{ForestGreen}{\# $L$: number of stages (network's depth) $||$ $N$: batch size}\\
\textcolor{ForestGreen}{\# $(H,W)$: input size $||$ $(H_l,W_l)$: feature size at stage $l$}\\
\textcolor{ForestGreen}{\# $\mathcal{E}^l$: encoder block (\ref{eq:encoder_features}--\ref{eq:transformer_block}) $||$ $\mathcal{D}^l$: $l$-th decoder block (\ref{eq:encoder_pcca}--\ref{eq:segmentation_output})}\\
\textcolor{ForestGreen}{\# $\mathrm{Up}$: upsampling op (bilinear or transposed conv)}\\
\textcolor{ForestGreen}{\# $\mathrm{Head}$: $1{\times}1$ conv + sigmoid (\ref{eq:segmentation_output})}\\
\textcolor{ForestGreen}{\# $\mathrm{PromptEncoder}$: (\ref{eq:coordnorm}--\ref{eq:promptscale}) $||$ $\mathbf{PCCA}$: (\ref{eq:avgpool}--\ref{eq:output})}\\[2pt]

\textcolor{ForestGreen}{\# \textbf{Mini-batch:} images $\mathbf{I}$, boxes $\mathbf{B}$, and labels $\mathbf{Y}$}

\For{ $\mathbf{I},\mathbf{B},\mathbf{Y}$ in \textbf{loader} }{
    \textcolor{ForestGreen}{\# \textbf{Prompt encoding (multi-scale)}}
    
    $\hat{\mathbf{F}}_b \leftarrow \mathrm{PromptEncoder}(\mathbf{B})$ \textcolor{ForestGreen}{\# (\ref{eq:coordnorm}--\ref{eq:convref}})\\
    $\{\mathbf{P}^l\}_{l=1}^{L} \leftarrow \mathrm{ResizeToPyramid}(\hat{\mathbf{F}}_b)$ \textcolor{ForestGreen}{\# (\ref{eq:promptscale})}

    \textcolor{ForestGreen}{\# \textbf{Encoder with PCCA and skip collection}}
    
    $\mathbf{x}^0 \leftarrow \mathrm{Stem}(\mathbf{I})$ \textcolor{ForestGreen}{\# optional pre-encoding conv}\\
    \textbf{skips} $\leftarrow$ [\ ]\\
    \For{$l=1$ \textbf{to} $L$}{
        $\mathbf{F}^{l} \leftarrow \mathcal{E}^l(\mathbf{x}^{l-1})$ \textcolor{ForestGreen}{\#~(\ref{eq:encoder_features} or \ref{eq:transformer_block})}\\
        $\hat{\mathbf{F}}^{l} \leftarrow \mathrm{PCCA}(\mathbf{F}^{l},\mathbf{P}^{l})$ \textcolor{ForestGreen}{\# (\ref{eq:encoder_pcca})}\\
        \textbf{append}(\textbf{skips}, $\hat{\mathbf{F}}^{l}$)\\
        $\mathbf{x}^{l} \leftarrow \mathrm{Down}(\hat{\mathbf{F}}^{l})$ \textcolor{ForestGreen}{\# stride-2 pooling/conv at $l{=}L$}
    }
    
    \textcolor{ForestGreen}{\# \textbf{Decoder with prompt re-injection via PCCA}}
    
    $\mathbf{G}^{L} \leftarrow \hat{\mathbf{F}}^{L}$ (bottleneck features)\\
    \For{$l=L$ \textbf{down to} $1$}{
        $\mathbf{U}^{l-1} \leftarrow \mathrm{Up}(\mathbf{G}^{l})$\\
        $\mathbf{S}^{l-1} \leftarrow \textbf{skips}[l-1]$ \textcolor{ForestGreen}{\#prompt-modulated skip}\\
        $\tilde{\mathbf{G}}^{l-1} \leftarrow \psi\big([\mathbf{U}^{l-1}\ \Vert\ \mathbf{S}^{l-1}]\big)$ \textcolor{ForestGreen}{\# Conv + BN + ReLU}\\
        $\hat{\mathbf{G}}^{l-1} \leftarrow \mathrm{PCCA}(\tilde{\mathbf{G}}^{l-1}, \mathbf{P}^{l-1})$ \textcolor{ForestGreen}{\# decoder prompting}\\
        $\mathbf{G}^{l-1} \leftarrow \hat{\mathbf{G}}^{l-1}$
    }

    \textcolor{ForestGreen}{\# \textbf{Prediction head and loss}}
    
    $\hat{\mathbf{S}} \leftarrow \mathrm{Head}(\mathbf{G}^{0})$ \textcolor{ForestGreen}{\# $1{\times}1$ conv + sigmoid,~\eqref{eq:segmentation_output}}\\
    $\mathcal{L} \leftarrow \mathrm{Loss}(\hat{\mathbf{S}}, \mathbf{Y})$ \textcolor{ForestGreen}{\# e.g., Dice + BCE,~\eqref{eq:total_loss}}

    \textcolor{ForestGreen}{\# \textbf{Backward/Update}}
    
    optimizer.zero\_grad()
    
    $\mathcal{L}.\mathrm{backward()}$
    
    optimizer.step()
}
\end{algorithm}

\begin{figure*}[!t]
\centering
\includegraphics[width=0.9\textwidth]{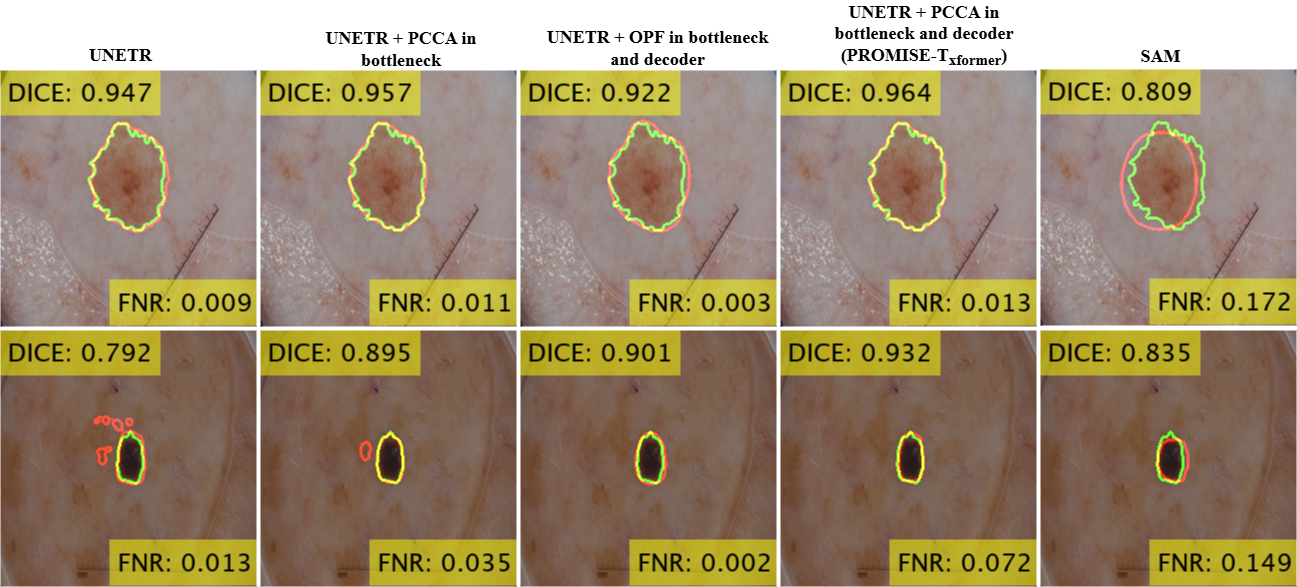}
\caption{Qualitative comparison of ISIC 2017 skin lesion segmentation. PROMISE-Net with hierarchical PCCA yields more complete and spatially coherent lesion delineations compared to UNETR and other baselines, demonstrating improved boundary precision and region completeness.}
\label{fig:ablation_visuals}
\end{figure*}

\section{Experimental Datasets and Settings}
\subsection{Datasets}

\paragraph{\textbf{ISIC-2017 Skin Lesion}}
This dataset~\cite{DBLP:journals/corr/abs-1710-05006} comprises 8-bit RGB dermoscopic images with spatial resolutions ranging from $540\times722$ to $4499\times6748$ pixels. It includes 2000 training, 150 validation, and 600 test images.

\paragraph{\textbf{Kvasir-Polyp}}
This dataset~\cite{jha2020kvasir} consists of 1000 high-quality colonoscopic frames with spatial resolutions ranging from $332\times352$ to $1920\times1072$ pixels. It is partitioned into 800 training, 100 validation, and 100 test images.

\paragraph{\textbf{Kvasir-Instrument}}
This dataset~\cite{10.1007/978-3-030-67835-7_19} serves as a benchmark for the segmentation of diagnostic and therapeutic instruments in gastrointestinal endoscopy, with spatial resolutions ranging from $571\times523$ to $1920\times1080$ pixels. It is partitioned into 472 training, 59 validation, and 59 test images.

\paragraph{\textbf{CAMUS-Cardiac}}
This dataset~\cite{8649738} comprises echocardiographic scans with heterogeneous image quality, with $35\%$ labeled as good, $46\%$ as medium, and $19\%$ as poor. It includes data from 400 patients for training, 50 for validation, and 50 for testing.

\subsection{Training Protocol and Evaluation}
\textbf{Experimental Settings:}  
All experiments were conducted using the PyTorch framework on an NVIDIA Tesla P100 GPU. Images were resized to $256 \times 256$ pixels. The models were optimized using the Adam optimizer with a learning rate of $1\times10^{-4}$ and trained for 150 epochs. Extensive offline data augmentation was applied to the CAMUS, Kvasir-Polyp, and Kvasir-Instrument datasets. Specifically, geometric transformations, including horizontal and vertical flips and random rotations of up to $360^\circ$, were employed. The augmented data were used during training and validation, while the test sets were kept untouched to ensure unbiased evaluation.

\textbf{Loss Function:}  
The overall training objective combines multi-class Dice loss with cross-entropy loss to jointly balance structural overlap and pixel-level classification accuracy \cite{hasan2020dsnet}:  
\begin{equation}
\mathcal{L}_{\text{total}} = \mathcal{L}_{\text{Dice}} + \lambda \, \mathcal{L}_{\text{CE}},
\label{eq:total_loss}
\end{equation}
where $\lambda = 0.5$ in all experiments.  
The Dice loss is averaged across all foreground classes to emphasize anatomical consistency, while the cross-entropy term stabilizes optimization in the multi-class segmentation setting.

\textbf{Evaluation:}  
Segmentation accuracy is evaluated using the Dice similarity coefficient (DSC) and Intersection-over-Union (IoU) for each class. Boundary agreement is quantified using the 95th percentile of the Hausdorff Distance (HD95), which measures the maximum symmetric surface deviation. To further measure pixel-level errors, we report the False Positive Rate (FPR) and False Negative Rate (FNR), capturing over- and under-segmentation, respectively. In addition to these spatial metrics, for cardiac datasets such as CAMUS, we assess the ejection fraction (EF)~\cite{hasan2025deep}, a clinically relevant functional biomarker defined as the ratio of stroke volume to end-diastolic volume (EDV). Agreement between estimated and reference EF values is analysed using Bland--Altman plots, which report the mean bias and limits of agreement (LoA) across the test cohort.

\section{Results and Discussion}
\subsection{PROMISENet Design and Ablation Analysis}
\label{subsec:ablation}
We perform ablation experiments to assess the impact of integrating the proposed PCCA module at different stages of PROMISE-T\textsubscript{xformer} on two datasets: ISIC-2017 and CAMUS-Cardiac. As summarized in Table~\ref{tab:TAM_SAM3}, these experiments quantify both segmentation performance and computational complexity when PCCA is inserted at the bottleneck alone or jointly at the bottleneck and decoder stages. For comparison, we also implement an alternative fusion scheme, termed outer product fusion (OPF), which replaces the global average pooling and shared latent projection in PCCA with a simple element-wise multiplication between image and prompt embeddings.

{\renewcommand{\arraystretch}{1.4} 
\begin{table*}[!ht]
\centering
\fontsize{8pt}{8pt}\selectfont
\caption{Ablation study of PCCA integration on the ISIC-2017 and CAMUS datasets. Progressive insertion of PCCA into deeper UNETR stages improves DSC and reduces FNR with modest computational overhead, demonstrating that hierarchical prompt-guided modulation enhances feature discrimination and robustness. Bold blue values denote the best performance.}
\label{tab:TAM_SAM3}
\begin{tabular}{lcc|cc|cc}
\toprule
\textbf{Methods} & \textbf{Params}($\downarrow$) & \textbf{FLOPs}($\downarrow$) &
\multicolumn{2}{c|}{\textbf{ISIC 2017}} &
\multicolumn{2}{c}{\textbf{CAMUS}} \\
\cmidrule(lr){4-5} \cmidrule(lr){6-7}
& & & \textbf{DSC} ($\%$) ($\uparrow$) & \textbf{FNR} ($\%$) ($\downarrow$) 
& \textbf{DSC} ($\%$) ($\uparrow$) & \textbf{FNR} ($\%$) ($\downarrow$) \\
\midrule
(1) UNETR \cite{hatamizadeh2022unetr}      & 116,048,130 & 52,792,357,120 & $82.5\pm18.0$  & $19.0\pm21.5$ & $88.6\pm5.4$  & $11.6\pm7.4$ \\
(2) UNETR + PCCA in bottleneck             & 127,922,994 & 58,231,075,072 & $88.3\pm9.6$   & $12.7\pm13.1$ & $89.8\pm4.7$  & $10.5\pm6.2$ \\
(3) UNETR + PCCA in bottleneck and decoder & 128,357,120 & 60,044,940,800 & \textcolor{blue}{$\boldsymbol{89.0\pm8.2}$}   & $10.6\pm10.9$ & \textcolor{blue}{$\boldsymbol{90.0\pm5.1}$}  & \textcolor{blue}{$\boldsymbol{9.2\pm5.8}$} \\
(4) UNETR + OPF in bottleneck and decoder  & 127,055,010 & 60,042,342,656 & $87.4\pm9.4$   & \textcolor{blue}{$\boldsymbol{10.2\pm11.4}$} & $89.6\pm5.0$  & $9.8\pm6.1$ \\
\hline
(1) vs. (2) ($\mathbf{p<0.05}$?) & & & \checkmark  & \checkmark & \checkmark  & \checkmark \\
(2) vs. (3) ($\mathbf{p<0.05}$?) & & & \checkmark  & \checkmark & \checkmark  & \checkmark \\
(3) vs. (4) ($\mathbf{p<0.05}$?) & & & \checkmark  & \xmark     & \checkmark  & \checkmark \\
\bottomrule
\end{tabular}
\end{table*}}

\begin{figure*}[!t]
\centering
\subfloat[Qualitative results on ISIC 2017 (\textcolor{green}{Green:} ground truth and \textcolor{red}{Red:} prediction).\label{fig:ablation_visuals}]{
\includegraphics[width=0.9\textwidth]{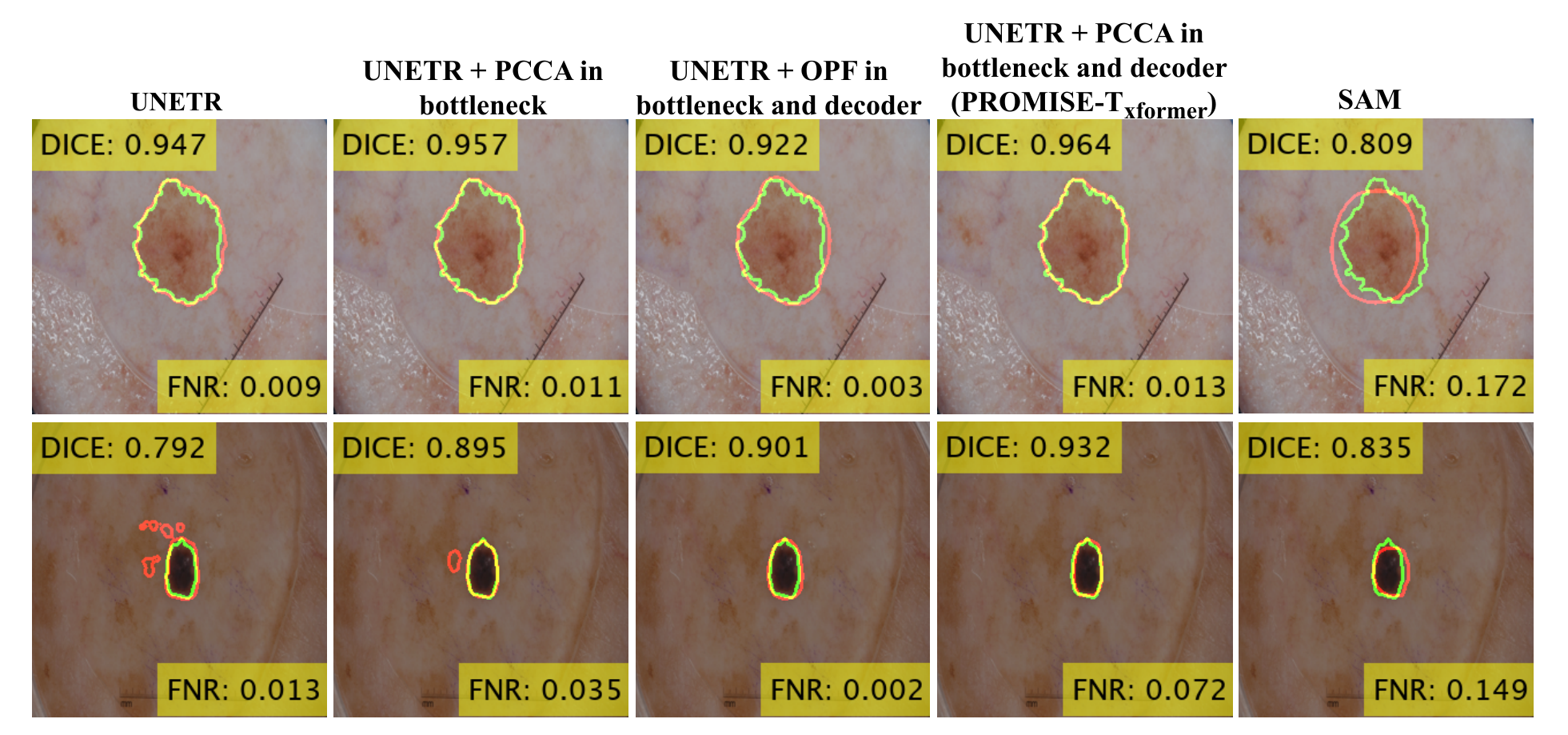}}\\
\subfloat[Intermediate feature-map comparison for UNETR variants, visualized by PCA projection to RGB space. \label{fig:Maps}]{
\includegraphics[width=0.9\textwidth, trim=20 20 15 9, clip]{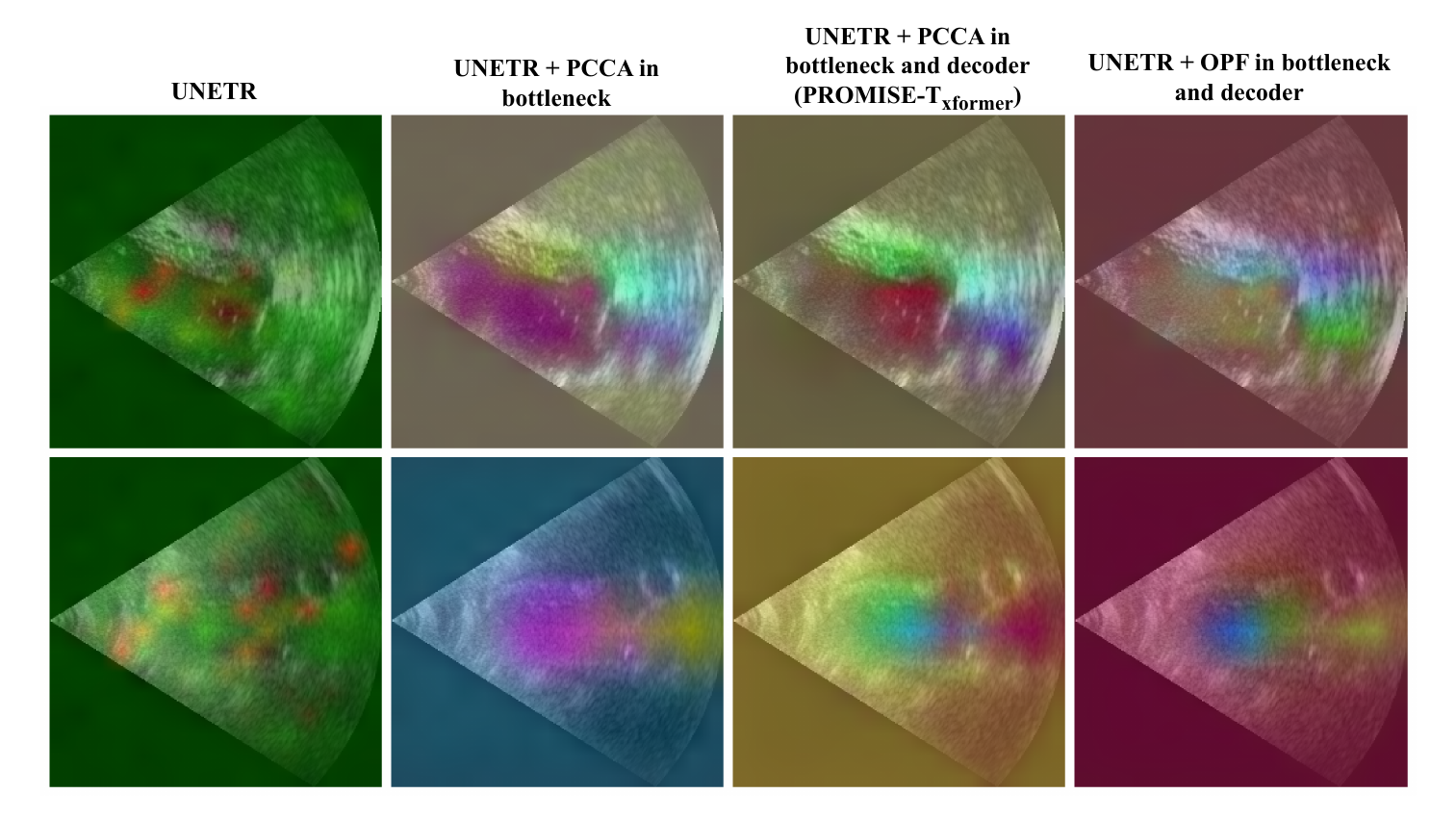}}
\caption{Ablation visualization in two-column layout: PROMISE-Net with hierarchical PCCA yields more complete delineations and more spatially focused feature representations than baseline UNETR and OPF fusion.}
\end{figure*}

\subsubsection{Impact of Hierarchical PCCA on Segmentation}
Table~\ref{tab:TAM_SAM3} demonstrates that integrating PCCA into the UNETR backbone consistently improves segmentation accuracy across both ISIC 2017 and CAMUS datasets. Incorporating PCCA at the bottleneck stage yields a significant improvement on ISIC 2017, raising DSC from 82.5 to 88.3 (+\textbf{5.8\%}, $p<0.05$) and reducing FNR from 19.0 to 12.7 (--\textbf{6.3\%}), while providing similar gains ($p<0.05$) on CAMUS (+\textbf{2.8\%} DSC and --\textbf{2.8\%} FNR). Extending PCCA to both the bottleneck and decoder stages further increases segmentation fidelity ($p < 0.05$), achieving a DSC of \textbf{89.0\%} and lowering the FNR to \textbf{10.6\%} on ISIC 2017, while attaining \textbf{92.1\%} DSC and \textbf{9.2\%} FNR on CAMUS.

In contrast, the simple OPF module with UNETR improves the DSC and FNR for both the datasets compared with the baseline UNETR (Table~\ref{tab:TAM_SAM3}), again confirming that introducing spatial prompts for feature modulation has a substantial impact on segmentation performance. However, when compared with the proposed PCCA, the OPF variant remains significantly inferior overall, as PCCA achieves higher overlap accuracy and stronger cross-dataset generalization. For ISIC 2017, the reduction in FNR achieved by OPF is not statistically significant relative to that of PCCA, indicating that while OPF captures some prompt-conditioned information, it lacks the robust hierarchical modulation achieved by PCCA. The parameter and FLOP differences between variants remain minimal ($<2\%$), confirming that the observed performance gains are primarily due to improved feature modulation by our novel PCCA rather than network scaling.

Visual comparisons in Fig.~\ref{fig:ablation_visuals} are consistent with the quantitative results in Table~\ref{tab:TAM_SAM3} showing PROMISE-T\textsubscript{xformer}'s better outputs with fewer discontinuities and reduced over-segmentation compared with the other baselines. Differences between methods are most apparent along lesion boundaries, where PROMISE-T\textsubscript{xformer} predictions more closely follow the ground truth contours. Fig.~\ref{fig:ablation_visuals} also includes results obtained using SAM. As reported in Table~\ref{tab:TAM_SAM}, SAM improves segmentation performance relative to the baseline UNETR; however, PROMISE-T\textsubscript{xformer} achieves higher overlap metrics than SAM on the evaluated datasets. These visual and quantitative comparisons indicate that the hierarchical integration of PCCA achieves the most balanced performance, enhancing segmentation robustness and boundary consistency with only marginal computational overhead.

\subsubsection{Hierarchical Modulation and Feature Separation}
Fig.~\ref{fig:Maps} provides qualitative insight into Table~\ref{tab:TAM_SAM3} on the CAMUS dataset, visualizing intermediate feature maps for UNETR and its PCCA variants. The baseline UNETR exhibits diffuse feature responses with substantial overlap between color components, indicating limited separation between anatomical regions. Introducing PCCA at the bottleneck results in more spatially concentrated feature activations and improved separation of color regions. When PCCA is applied at both the bottleneck and decoder stages (PROMISE-T\textsubscript{xformer}), corresponding to the highest quantitative performance on CAMUS (DSC 90.0\%, FNR 9.2\%), the feature embeddings appear more localized and exhibit clearer separation across distinct color components (Fig.~\ref{fig:Maps}). In contrast, the OPF-based variant shows less consistent separation of feature responses compared with the hierarchical PCCA configuration. These visual patterns align with the quantitative trends in Table \ref{tab:TAM_SAM3}, where hierarchical PCCA integration yields higher overlap accuracy and lower false-negative rates than OPF.

\subsubsection{Robustness to Channel Compression in PCCA}
The reduction ratio $r$ in PCCA controls the level of channel compression in the excitation MLP. Table~\ref{tab:r_ablation} reports segmentation performance and model complexity on the ISIC 2017 dataset for $r \in \{2, 4, 8, 16, 32\}$. Across all configurations, HD95 varies within a narrow range (12.7--12.9), and FNR remains between 10.3\% and 11.2\%, indicating limited sensitivity to the choice of $r$. Among the evaluated settings, $r=16$ yields the lowest FNR (10.3\%) and a low HD95 (12.8) while maintaining moderate computational cost (52.15M parameters and 106.33G FLOPs). Parameter count and FLOPs decrease monotonically with increasing $r$, with differences remaining small across configurations. Based on these, $r=16$ is used in all experiments.

\begin{table}[!t]
\centering
\caption{Effect of the reduction ratio \(r\) on PCCA efficiency and performance on ISIC-2017. A moderate setting (\(r{=}16\)) offers the best accuracy--efficiency trade-off, achieving low HD95 and FNR with minimal overhead. Bold blue indicates the best result.}
\label{tab:r_ablation}
\setlength{\tabcolsep}{3.5pt}
\small
\resizebox{\columnwidth}{!}{%
\begin{tabular}{lcccc}
\toprule
\textbf{Reduction Ratio ($r$)} & \textbf{HD95} ($\downarrow$) & \textbf{FNR} ($\%$) ($\downarrow$) & \textbf{FLOPs} ($\downarrow$) & \textbf{Params} ($\downarrow$) \\
\midrule
$r=2$   & 12.9 & 10.5 & 106.33347 & 53,070,338 \\
$r=4$   & 12.9 & 11.2 & 106.33242 & 52,545,794 \\
$r=8$   & 12.9 & 10.9 & 106.33190 & 52,283,522 \\
$r=16$  & \textcolor{blue}{$\boldsymbol{12.8}$} & \textcolor{blue}{$\boldsymbol{10.3}$} & 106.33164 & 52,152,386 \\
$r=32$  & 12.7 & 10.6 & 106.33150 & 52,086,818 \\
\bottomrule
\end{tabular}
}
\end{table}

\subsection{Generalizability Analysis}
\label{subsec:cross_modal}
{\renewcommand{\arraystretch}{1.4} 
\begin{table*}[!t]
\begin{center}
\fontsize{10pt}{10pt}\selectfont
\caption{Performance comparison across four segmentation datasets: ISIC-2017 (skin lesion), Kvasir-Polyp, Kvasir-Instrument (endoscopic tool), and CAMUS (cardiac echocardiography). PROMISE-Net consistently outperforms baseline architectures across diverse anatomical structures and imaging modalities, demonstrating strong cross-domain and cross-architectural generalizability. Bold blue values indicate the best performance for each metric.}
\label{tab:TAM_SAM}
\begin{tabular}{lcccc}
\toprule
\textbf{Methods} & \textbf{DSC} ($\%$) ($\uparrow$) & \textbf{IoU} ($\%$) ($\uparrow$) & \textbf{HD95 (Pix)} ($\downarrow$) & \textbf{FNR} ($\%$) ($\downarrow$) \\
\midrule
\multicolumn{5}{c}{\textbf{Skin Lesion Dataset (ISIC-2017)} \cite{DBLP:journals/corr/abs-1710-05006}} \\
\midrule
(1) U-Net \cite{ronneberger2015u} & $82.2\pm 18.8$ & $73.3\pm 21.8$ & $23.6\pm 26.4$ & $19.4\pm 21.9$ \\
(2) PROMISE-C\textsubscript{NN} (proposed)& \textcolor{blue}{$\boldsymbol{90.7\pm 7.5}$} & \textcolor{blue}{$\boldsymbol{83.7\pm 10.0}$} & \textcolor{blue}{$\boldsymbol{11.7\pm 10.6}$} & \textcolor{blue}{$\boldsymbol{7.3\pm 9.0}$} \\ 
(1) vs. (2) ($\mathbf{p<0.05}$?) & \checkmark  & \checkmark & \checkmark & \checkmark \\ \hline
(3) UNETR \cite{hatamizadeh2022unetr} & $82.5\pm 18.0$ & $73.4\pm 21.4$ & $22.3\pm 25.2$ & $19.0\pm 21.5$ \\
(4) PROMISE-T\textsubscript{xformer} (proposed)& $89.0\pm 8.2$ & $81.0\pm 11.4$ & $15.7\pm 18.2$ & $10.6\pm 10.9$ \\
(3) vs. (4) ($\mathbf{p<0.05}$?) & \checkmark  & \checkmark & \checkmark & \checkmark \\ \hline
(5) SAM \cite{kirillov2023segment} & $85.1\pm 9.0$ & $75.0\pm 11.5$ & $16.7\pm 10.7$ & $12.1\pm 9.1$ \\
(2) vs. (5) ($\mathbf{p<0.05}$?) & \checkmark  & \checkmark & \checkmark & \checkmark \\ 
(4) vs. (5) ($\mathbf{p<0.05}$?) & \checkmark  & \checkmark & \checkmark & \checkmark \\
\midrule
\multicolumn{5}{c}{\textbf{Polyp Segmentation Dataset (Kvasir-Polyp)} \cite{jha2020kvasir}} \\
\midrule
(1) U-Net \cite{ronneberger2015u} & $86.0\pm 17.3$ & $78.5\pm 20.8$ & $29.1\pm 33.8$ & $11.9\pm 19.7$ \\
(2) PROMISE-C\textsubscript{NN} (proposed) & \textcolor{blue}{$\boldsymbol{92.5\pm 9.8}$} & \textcolor{blue}{$\boldsymbol{87.2\pm 13.4}$} & \textcolor{blue}{$\boldsymbol{13.3\pm 18.6}$} & \textcolor{blue}{$\boldsymbol{7.3\pm 11.4}$} \\
(1) vs. (2) ($\mathbf{p<0.05}$?) & \checkmark  & \checkmark & \checkmark & \checkmark \\ \hline
(3) UNETR \cite{hatamizadeh2022unetr} & $68.3\pm 25.7$ & $57.0\pm 27.0$ & $58.0\pm 37.8$ & $23.6\pm 27.7$ \\
(4) PROMISE-T\textsubscript{xformer} (proposed) & $87.5\pm 14.8$ & $80.0\pm 17.9$ & $23.2\pm 33.9$ & $9.0\pm 13.1$ \\
(3) vs. (4) ($\mathbf{p<0.05}$?) & \checkmark  & \checkmark & \checkmark & \checkmark \\ \hline
(5) SAM \cite{kirillov2023segment} & $80.0\pm 13.1$ & $68.2\pm 15.4$ & $24.4\pm 24.0$ & $14.7\pm 10.3$ \\
(2) vs. (5) ($\mathbf{p<0.05}$?) & \checkmark  & \checkmark & \checkmark & \checkmark \\ 
(4) vs. (5) ($\mathbf{p<0.05}$?) & \checkmark  & \checkmark & \xmark & \checkmark \\
\midrule
\multicolumn{5}{c}{\textbf{Diagnostic and Therapeutic Tool Segmentation (Kvasir-Instrument)} \cite{10.1007/978-3-030-67835-7_19}} \\
\midrule
(1) U-Net \cite{ronneberger2015u} & $94.4\pm 13.8$ & $91.2\pm 14.8$ & $8.7\pm 19.5$ & $6.0\pm 15.3$ \\
(2) PROMISE-C\textsubscript{NN} (proposed) & \textcolor{blue}{$\boldsymbol{97.2\pm 1.4}$} & \textcolor{blue}{$\boldsymbol{94.6\pm 2.7}$} & \textcolor{blue}{$\boldsymbol{3.4\pm 3.0}$} & \textcolor{blue}{$\boldsymbol{2.7\pm 1.8}$} \\
(1) vs. (2) ($\mathbf{p<0.05}$?) & \xmark  & \checkmark & \checkmark & \checkmark \\ \hline
(3) UNETR \cite{hatamizadeh2022unetr} & $95.3\pm 7.2$ & $91.6\pm 9.8$ & $8.1\pm 22.0$ & $4.1\pm 6.9$ \\
(4) PROMISE-T\textsubscript{xformer} (proposed) & $96.2\pm 2.2$ & $92.7\pm 4.0$ & $4.4\pm 2.5$ & $4.5\pm 4.0$ \\
(3) vs. (4) ($\mathbf{p<0.05}$?) & \xmark  & \xmark & \xmark & \xmark \\
\midrule
\multicolumn{5}{c}{\textbf{2D Echocardiography (CAMUS)} \cite{8649738}} \\
\midrule
(1) U-Net \cite{ronneberger2015u} & $90.8\pm 4.8$ & $83.6\pm 7.3$ & $6.2\pm 4.1$ & $9.4\pm 6.6$ \\
(2) PROMISE-C\textsubscript{NN} (proposed) & \textcolor{blue}{$\boldsymbol{91.3\pm 4.1}$} & \textcolor{blue}{$\boldsymbol{84.4\pm 6.4}$} & \textcolor{blue}{$\boldsymbol{5.9\pm 4.1}$} & \textcolor{blue}{$\boldsymbol{8.4\pm 5.2}$} \\
(1) vs. (2) ($\mathbf{p<0.05}$?) & \checkmark  & \checkmark & \xmark & \checkmark \\ \hline
(3) UNETR \cite{hatamizadeh2022unetr} & $88.6\pm 5.4$ & $80.1\pm 8.1$ & $7.9\pm 5.2$ & $11.6\pm 7.4$ \\
(4) PROMISE-T\textsubscript{xformer} (proposed) & $90.0\pm 5.1$ & $82.2\pm 7.5$ & $7.0\pm 5.2$ & $9.2\pm 5.8$ \\
(3) vs. (4) ($\mathbf{p<0.05}$?) & \checkmark  & \checkmark & \checkmark & \checkmark \\
\bottomrule
\end{tabular}
\end{center}
\end{table*}
}

\subsubsection{Cross-Architectural Generalization}
Cross-architectural generalization is evaluated by PROMISE-C\textsubscript{NN} and PROMISE-T\textsubscript{xformer} backbones and comparing each against its corresponding baseline U-Net and UNETR across four datasets (Table~\ref{tab:TAM_SAM}). 

\begin{figure*}[!ht]
\centering

\begin{minipage}{0.495\textwidth}
\centering
\subfloat[Skin lesion segmentation on ISIC 2017]{
\includegraphics[width=\linewidth]{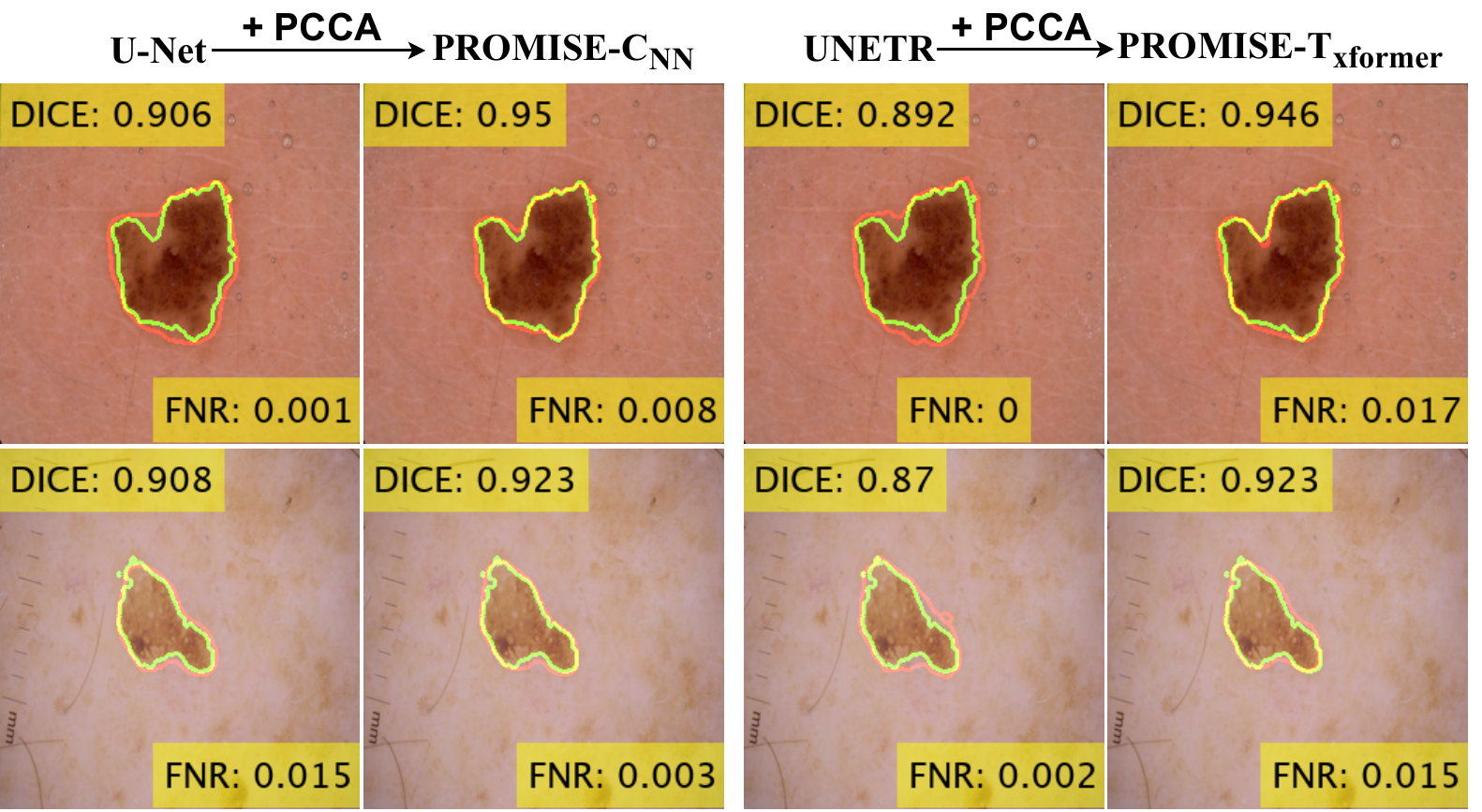}}
\end{minipage}
\begin{minipage}{0.495\textwidth}
\centering
\subfloat[Diagnostic and therapeutic instrument segmentation on Kvasir-instrument]{
\includegraphics[width=\linewidth]{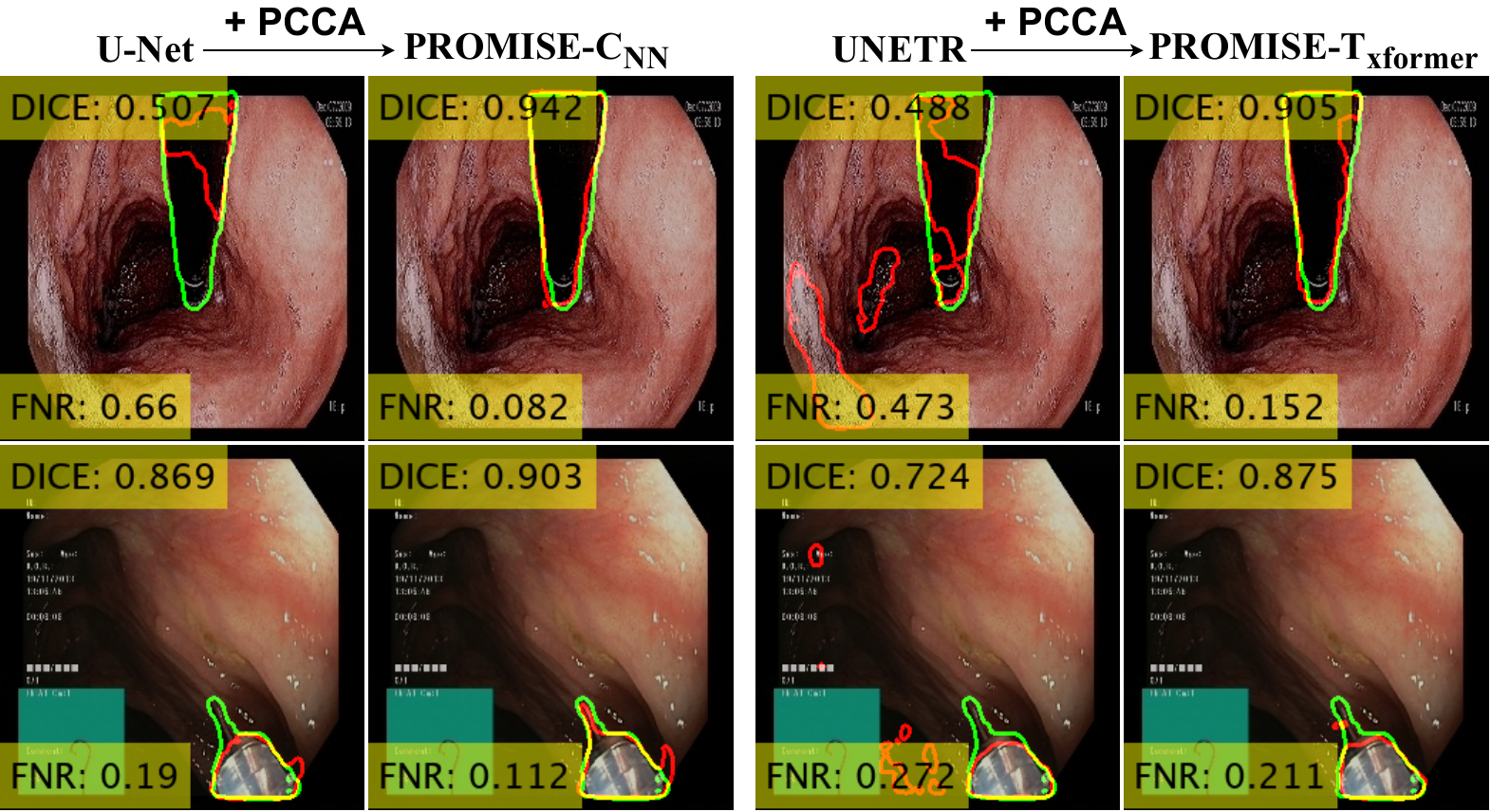}}
\end{minipage}

\begin{minipage}{0.495\textwidth}
\centering
\subfloat[Gastrointestinal polyp segmentation on Kvasir-Polyp]{
\includegraphics[width=\linewidth]{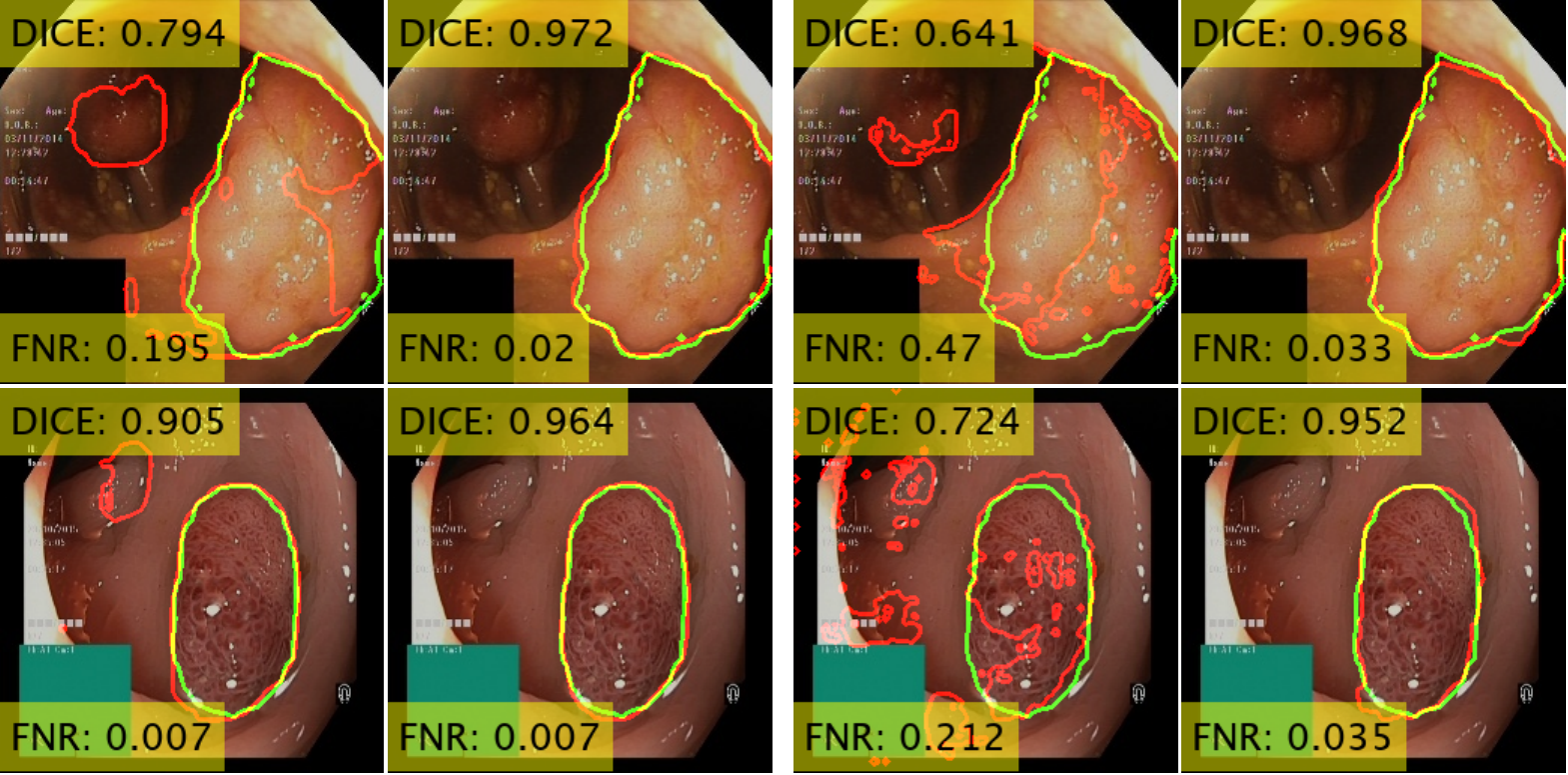}}
\end{minipage}
\begin{minipage}{0.495\textwidth}
\centering
\subfloat[Multiclass cardiac chamber segmentation on CAMUS]{
\includegraphics[width=\linewidth]{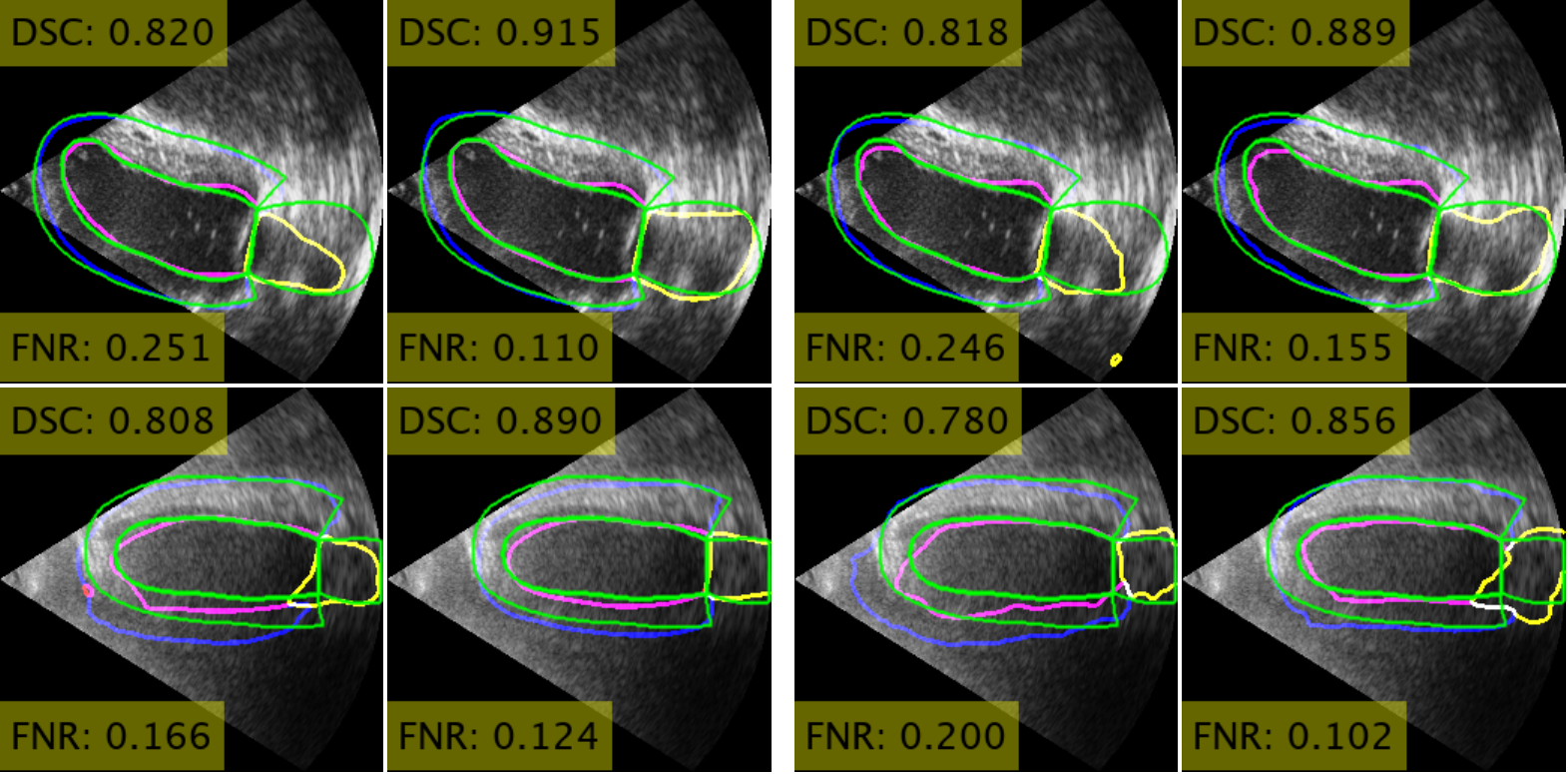}}
\end{minipage}
\caption{Comparison of baseline segmentation networks (U-Net and UNETR) and their PCCA-integrated counterparts, PROMISE-C\textsubscript{NN} and PROMISE-T\textsubscript{xformer}, across cross-modal and cross-anatomical datasets. In (a)–(c), ground-truth boundaries are shown in green and predictions in red for binary segmentation tasks. In (d), for multi-class cardiac segmentation on CAMUS, green contours denote ground truth, while blue, purple, and yellow indicate predicted MYO, LV, and LA boundaries, respectively. PROMISE-Net variants exhibit improved anatomical fidelity, smoother contours, and reduced fragmentation compared to their baselines.}
\label{fig:Qualitative_4Datasets}
\end{figure*}

For convolutional networks, PROMISE-C\textsubscript{NN} improves DSC over U-Net by \textbf{+8.5\%} on ISIC 2017 (82.2\% to 90.7\%), \textbf{+6.5\%} on Kvasir-Polyp (86.0\% to 92.5\%), \textbf{+2.8\%} on Kvasir-Instrument (94.4\% to 97.2\%), and \textbf{+0.5\%} on CAMUS (90.8\% to 91.3\%). These gains are accompanied by substantial reductions in boundary error, including approximately \textbf{50\%} on ISIC 2017 (23.6 to 11.7 pixels) and \textbf{55\%} on Kvasir-Polyp (29.1 to 13.3 pixels), as well as consistent decreases in false-negative rate across all datasets. Qualitative examples in Fig.~\ref{fig:Qualitative_4Datasets} (a,c) corroborate these trends, showing more complete delineations with reduced fragmentation relative to U-Net.

Similarly, PROMISE-T\textsubscript{xformer} consistently outperforms UNETR across all evaluated datasets. On ISIC 2017 and CAMUS, DSC increases from 82.5\% to 89.0\% and from 88.6\% to 90.0\%, respectively, with corresponding reductions in FNR. On Kvasir-Polyp and Kvasir-Instrument, PROMISE-T\textsubscript{xformer} achieves higher DSC and lower HD95 than UNETR, with statistically significant differences reported for most metrics. Visual comparisons in Fig.~\ref{fig:Qualitative_4Datasets} (b,d) further show smoother contours and more continuous anatomical structures. 

Together, the consistent quantitative gains across both convolutional and transformer-based backbones, supported by corresponding qualitative evidence, indicate that the performance gains introduced by PCCA are not specific to a particular network type.

\subsubsection{Cross-Anatomical and Cross-Domain Generalization}
Cross-anatomical generalization is assessed across segmentation tasks involving distinct anatomical targets (Table~\ref{tab:TAM_SAM}), including skin lesions (ISIC 2017), gastrointestinal polyps (Kvasir-Polyp), surgical instruments (Kvasir-Instrument), and cardiac chambers (CAMUS). 

On ISIC 2017, PROMISE-C\textsubscript{NN} improves DSC from 82.2\% to 90.7\%, reduces HD95 from 23.6 to 11.7 pixels, and lowers FNR from 19.4\% to 7.3\%. On Kvasir-Polyp, DSC increases from 86.0\% to 92.5\%, with HD95 reduced from 29.1 to 13.3 pixels and FNR from 11.9\% to 7.3\%. For the anatomically distinct Kvasir-Instrument dataset, PROMISE-C\textsubscript{NN} improves DSC from 94.4\% to 97.2\%, reduces HD95 from 8.7 to 3.4 pixels, and lowers FNR from 6.0\% to 2.7\%. On the CAMUS dataset, DSC increases from 90.8\% to 91.3\%, with a corresponding reduction in FNR (Table~\ref{tab:TAM_SAM}). Qualitative results in Fig.~\ref{fig:Qualitative_4Datasets} align with these quantitative improvements. PROMISE-Net variants produce lesion and polyp segmentations with fewer disconnected regions and closer adherence to annotated boundaries on ISIC 2017 and Kvasir-Polyp. In Kvasir-Instrument, PROMISE-Net more completely delineates elongated tool structures in regions affected by strong specular reflections, where baseline models exhibit partial detections or discontinuities. On CAMUS, PROMISE-Net yields smoother myocardial contours and clearer separation between cardiac chambers, consistent with the improved overlap accuracy and reduced false-negative rates reported in Table~\ref{tab:TAM_SAM}.

These results show that the prompt-conditioned modulation in PROMISE-Net transfers effectively across anatomically diverse targets without task-specific architectural modifications.

{\renewcommand{\arraystretch}{1.4}
\begin{table*}[!ht]
\centering
\fontsize{9pt}{8pt}\selectfont
\caption{%
Quantitative comparison of the proposed PROMISE-C\textsubscript{NN} with state-of-the-art segmentation methods across four datasets:
ISIC-2017 (skin lesion), Kvasir-Polyp, Kvasir-Instrument (endoscopic tool), and CAMUS (2D echocardiography). Result on CAMUS is shown for the LA class.
Bold blue values indicate the best performance for each metric.
}
\label{tab:TAM_SAM4}

\begin{minipage}[t]{0.48\textwidth}
\centering
\textbf{(a) ISIC-2017 (Skin Lesion)}\\[2pt]
\begin{tabular}{lccc}
\toprule
Methods & DSC ($\%$) ($\uparrow$) & IoU ($\%$) ($\uparrow$)  & FNR ($\%$) ($\downarrow$) \\
\midrule
Pact-Net \cite{chen2023pact} & 86.2 & 79.3  & 13.8 \\
USL-Net \cite{li2024usl} & 80.5 & 68.5  & 11.4 \\
FAT-Net \cite{wu2022fat} & 85.0 & 76.5 & 16.1 \\
HTC-Net \cite{tang2024htc} & 90.1 & \textcolor{blue}{\textbf{84.0}}  & 11.8 \\
\textbf{PROMISE-C\textsubscript{NN}} & \textcolor{blue}{\textbf{90.7}} & 83.7  & \textcolor{blue}{\textbf{7.3}} \\
\bottomrule
\end{tabular}
\end{minipage}
\hfill
\begin{minipage}[t]{0.48\textwidth}
\centering
\textbf{(b) Kvasir-Polyp}\\[2pt]
\begin{tabular}{lccc}
\toprule
Methods & DSC ($\%$) ($\uparrow$) & IoU ($\%$) ($\uparrow$)  & FNR ($\%$) ($\downarrow$) \\
\midrule
RM-UNet \cite{tang2024rm} & 90.2 & 82.2  & 10.3 \\
Pact-Net \cite{chen2023pact} & 90.6 & 84.7  & -- \\
USL-Net \cite{li2024usl} & 91.2 & 85.9  & -- \\
ERDUNet \cite{li2023erdunet} & 90.7 & 84.6 &  -- \\
\textbf{PROMISE-C\textsubscript{NN}} & \textcolor{blue}{\textbf{92.5}} & \textcolor{blue}{\textbf{87.2}}  & \textcolor{blue}{\textbf{7.3}} \\
\bottomrule
\end{tabular}
\end{minipage}

\vspace{6pt}

\begin{minipage}[t]{0.48\textwidth}
\centering
\textbf{(c) Kvasir-Instrument}\\[2pt]
\begin{tabular}{lcccc}
\toprule
Methods & DSC ($\%$) ($\uparrow$) & IoU ($\%$) ($\uparrow$)  & FNR ($\%$) ($\downarrow$) \\
\midrule
RM-UNet \cite{tang2024rm} & 94.7 & 89.9  & 5.9 \\
MAF-Net \cite{yang2023maf} & 96.8 & \textcolor{blue}{\textbf{96.6}}  & -- \\
DECA-Net \cite{liang2024deca} & 96.9 & 93.9 & -- \\
ERDUNet \cite{li2023erdunet} & 95.2 & 91.6  & -- \\
\textbf{PROMISE-C\textsubscript{NN}} & \textcolor{blue}{\textbf{97.2}} & 94.6  & \textcolor{blue}{\textbf{2.7}} \\
\bottomrule
\end{tabular}
\end{minipage}
\hfill
\begin{minipage}[t]{0.48\textwidth}
\centering
\textbf{(d) CAMUS (2D Echocardiography)}\\[2pt]
\begin{tabular}{lccc}
\toprule
Methods & DSC ($\%$) ($\uparrow$) & IoU ($\%$) ($\uparrow$) & HD95 ($mm$) ($\downarrow$)  \\
\midrule
CLAS \cite{10.1007/978-3-030-59713-9_60} & 91.4 & -- & 5  \\
EchoSAM \cite{LI2025108000} & 90.7 &  83.4 & 3.8  \\
TAM-FCN8s \cite{hasan2025motion} & 91.6 & -- & \textcolor{blue}{\textbf{3.3}}  \\
CoST-UNet \cite{ISLAM2024106633} & 87.6 & 79.2 & 6.7  \\
\textbf{PROMISE-C\textsubscript{NN}} & \textcolor{blue}{\textbf{91.8}} & \textcolor{blue}{\textbf{85.2}} & 3.9  \\
\bottomrule
\end{tabular}
\end{minipage}

\end{table*}
}

\subsubsection{Cross Image Quality Levels and Cardiac Structures}
\begin{figure}[!htbp]
\centering
\includegraphics[width=\linewidth, keepaspectratio]{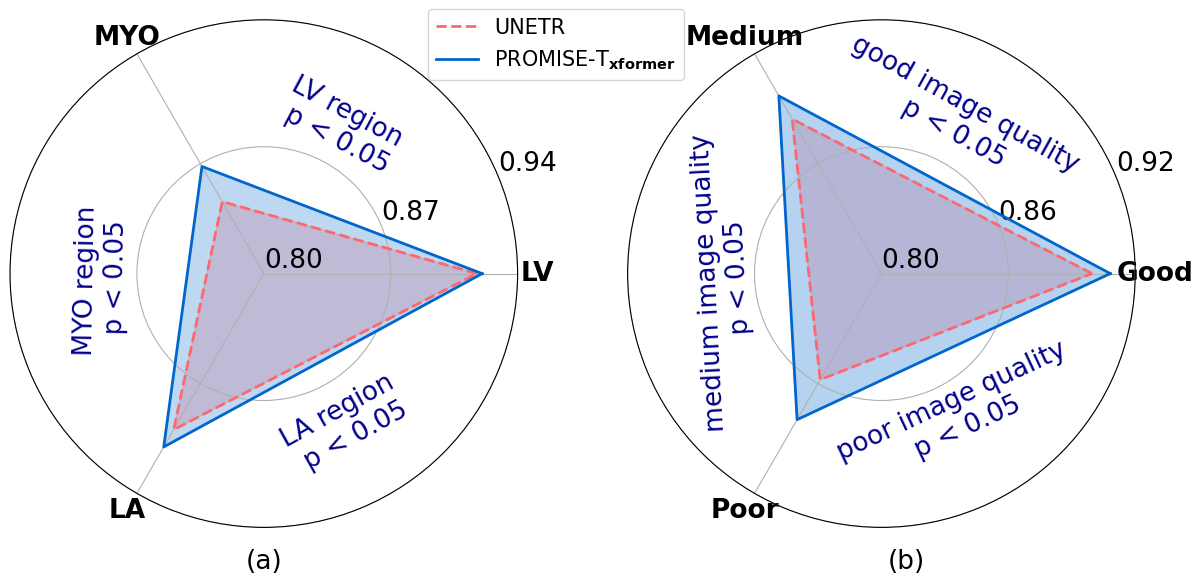}
\caption{Comparison of segmentation performance between UNETR and PROMISE-T\textsubscript{xformer} on CAMUS: class-wise DSC (LV, MYO, and LA) and DSC across good, medium, and poor image quality levels \cite{8649738}.}
\label{fig:dsc_radar}
\end{figure}

Image quality variation robustness is evaluated on the CAMUS dataset by stratifying test samples into good, medium, and poor quality categories, as defined in the dataset annotations \cite{8649738}. As shown in Fig.~\ref{fig:dsc_radar}, PROMISE-T\textsubscript{xformer} achieves higher DSC than UNETR across all levels. The variation in DSC between good and poor quality images for PROMISE-T\textsubscript{xformer} remains within 5\%, indicating limited sensitivity to image quality degradation relative to the baseline.

Class-wise analysis in Fig.~\ref{fig:dsc_radar} further demonstrates consistent improvements in DSC for all cardiac structures, including LV, MYO, and LA, with statistically significant differences reported for each class. Together with the binary segmentation results on ISIC-2017 and Kvasir datasets, these findings indicate that PROMISE-Net generalizes across label-space complexity, maintaining performance gains when transitioning from binary to multi-class segmentation tasks. Overall, the performance gains of PROMISE-T\textsubscript{xformer} are preserved across both varying image quality conditions and multiple cardiac regions on the CAMUS dataset.

\subsection{PCCA vs. Alternative SAM in Performance and Efficiency}
PROMISE-Net is further compared with the alternative prompt-based SAM on the ISIC 2017 and Kvasir-Polyp datasets in Table~\ref{tab:TAM_SAM}. On ISIC-2017, PROMISE-C\textsubscript{NN} achieves a DSC of 90.7\% compared with 85.1\% for SAM, while reducing HD95 from 16.7 to 11.7 pixels and FNR from 12.1\% to 7.3\%. On Kvasir-Polyp, PROMISE-C\textsubscript{NN} attains a DSC of 92.5\% versus 80.0\% for SAM, with HD95 reduced from 24.4 to 13.3 pixels and FNR from 14.7\% to 7.3\%. PROMISE-T\textsubscript{xformer} similarly outperforms SAM in terms of DSC and FNR on both datasets, with statistically significant differences reported for most metrics (Table~\ref{tab:TAM_SAM}). These quantitative differences are consistent with the qualitative comparisons in Fig.~\ref{fig:ablation_visuals}, where PROMISE-Net variants produce more complete and spatially coherent segmentations than SAM, indicating that hierarchical prompt-conditioned modulation provides advantages over late-stage prompt integration alone.

Again, as shown in Table~\ref{tab:model_efficiency}, PROMISE-C\textsubscript{NN} attains 90.7\% DSC with 118.15~GFLOPs and 61~million parameters, outperforming U-Net (82.2\%, 96.6~GFLOPs, 31~M) and remaining far lighter than SAM (85.1\%, 2991.32~GFLOPs, 312~M). This balance of accuracy and computational cost makes PROMISE-Net suitable for practical deployment across diverse medical imaging modalities.

{\renewcommand{\arraystretch}{1}
\begin{table}[!t]
\centering
\caption{Computational efficiency, model complexity, and performance comparison of selected segmentation architectures. DSC is reported on ISIC-2017; bold blue denotes the best result per metric.}
\label{tab:model_efficiency}
\begin{tabular}{lccc}
\toprule
\textbf{Model} & \textbf{DSC} & \textbf{FLOPs} & \textbf{Params} \\
\midrule
U-Net & $82.2$ & $96.6$ & $31$ \\
SAM & $85.1$ & $2991.32$ & $312$ \\
\textbf{PROMISE-C\textsubscript{NN}} & \textcolor{blue}{$\boldsymbol{90.7}$} & \textcolor{blue}{$\boldsymbol{118.15}$} & \textcolor{blue}{$\boldsymbol{61}$} \\
\bottomrule
\end{tabular}
\end{table}
}

\subsection{Robustness to User Prompt and Interaction Variability}
\begin{figure}[!htbp]
\centering
\includegraphics[width=\linewidth]{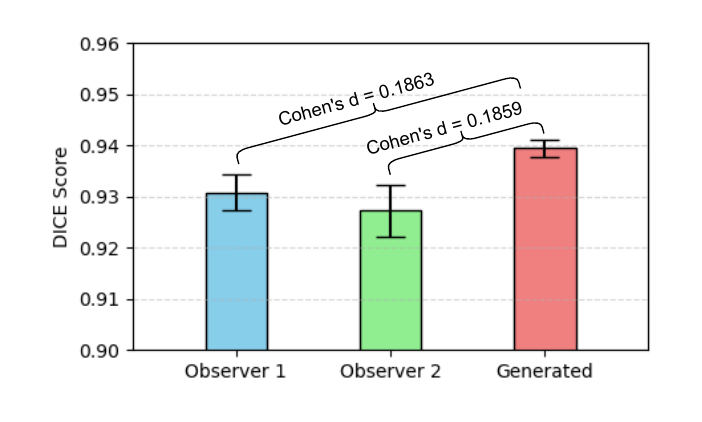}
\caption{Dice score comparison using observer-drawn and automatically generated bounding-box prompts, with pairwise effect sizes (Cohen’s $d$).}
\label{fig:observer_dice_cohensd}
\end{figure}

To assess robustness to interaction variability, bounding-box prompts from two human observers were compared with automatically generated boxes on 300 ISIC-2017 test images. To simulate realistic user imprecision, ground-truth bounding boxes were synthetically perturbed by $\pm20$ pixels. PROMISE-Net exhibits comparable performance across observer-provided and automated prompts; pairwise effect sizes between observer-driven and automated prompts remain small (Cohen’s $d < 0.2$), indicating negligible practical differences in the results (Fig.~\ref{fig:observer_dice_cohensd}). These results demonstrate that PROMISE-Net is robust to inter-observer variability and moderate prompt perturbations, producing stable and reproducible segmentations despite differences in prompt placement or annotation style.

\begin{figure*}[!t]
\centering
\subfloat[EDV of UNETR]{\includegraphics[width=0.31\textwidth]{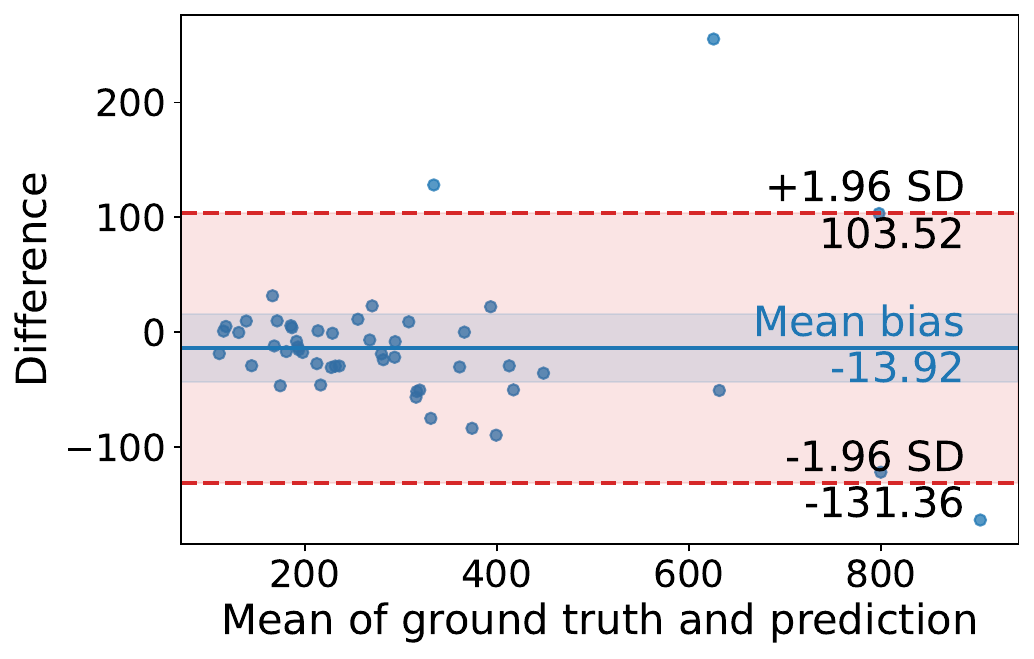}}\hfill
\subfloat[ESV of UNETR]{\includegraphics[width=0.31\textwidth]{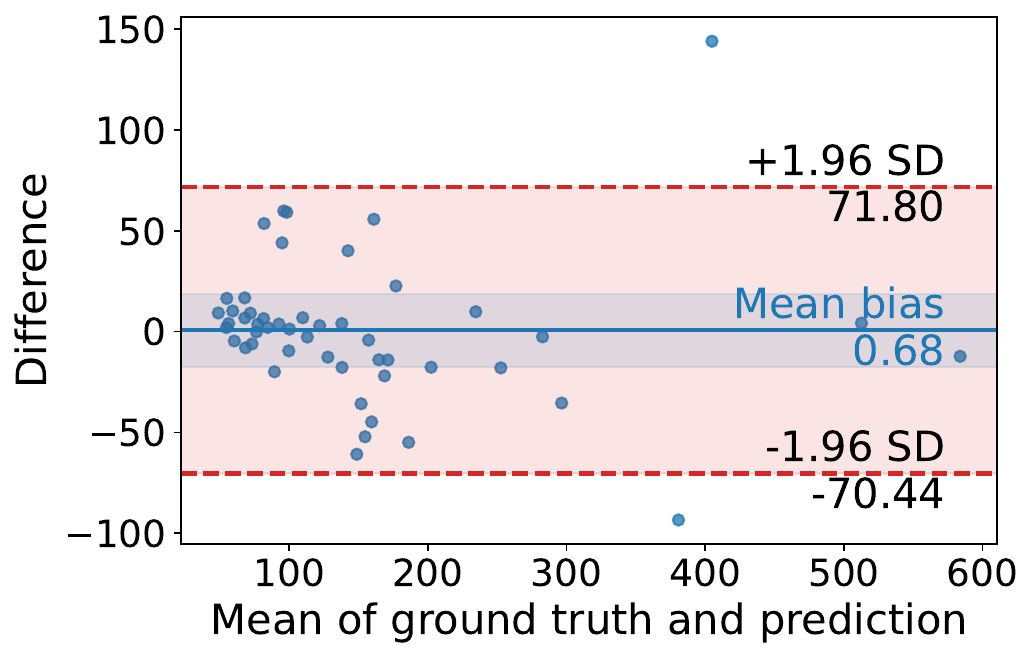}}\hfill
\subfloat[EF of UNETR]{\includegraphics[width=0.31\textwidth]{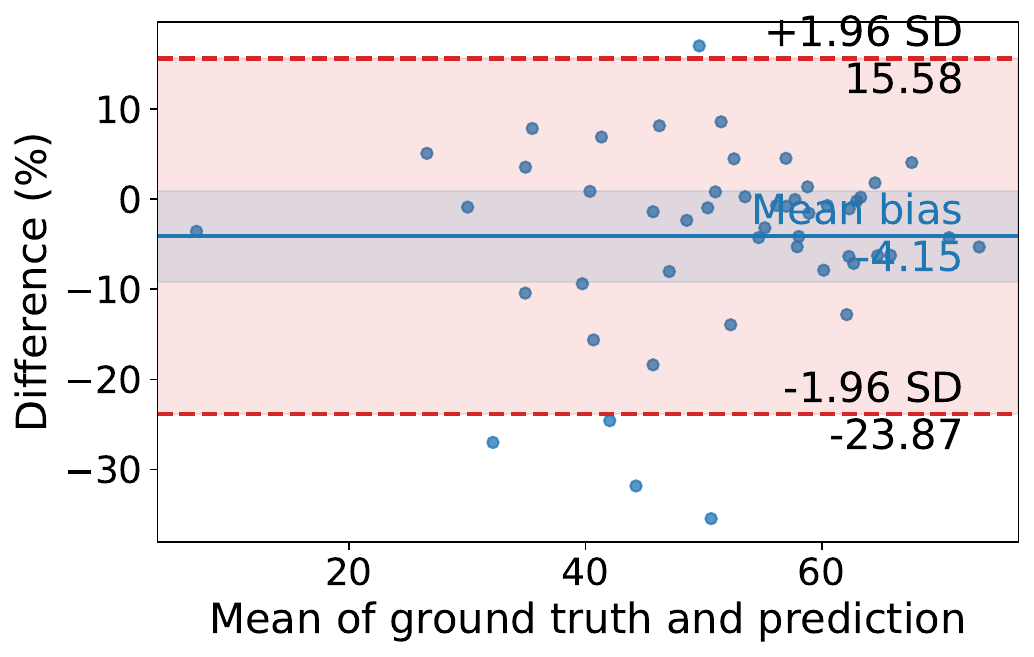}}\\[3pt]
\subfloat[EDV of our PROMISE-T\textsubscript{xformer}]{\includegraphics[width=0.31\textwidth]{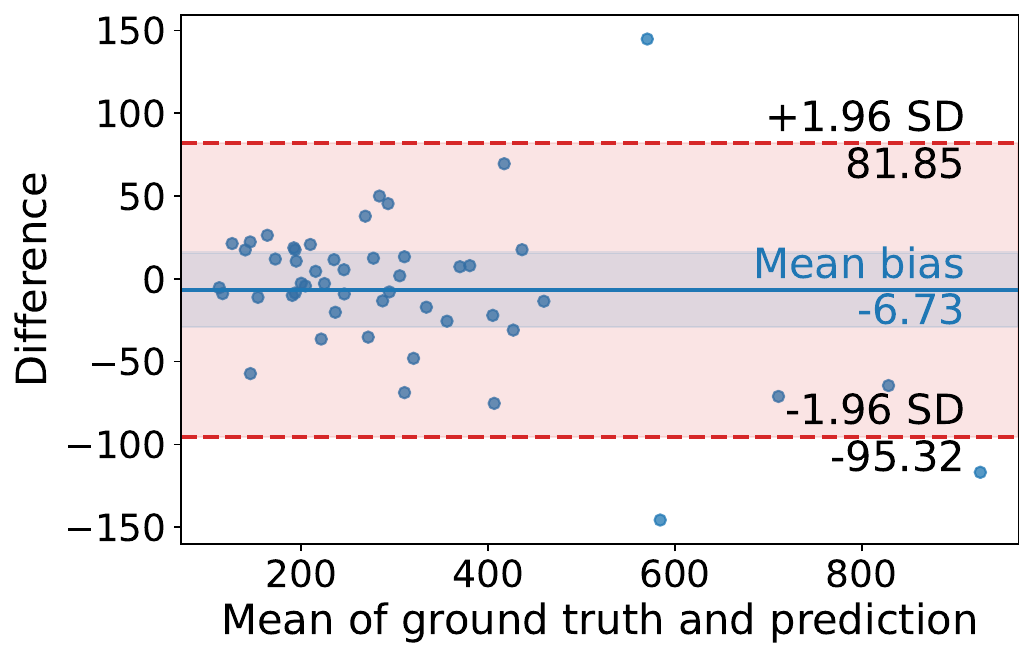}}\hfill
\subfloat[ESV of our PROMISE-T\textsubscript{xformer}]{\includegraphics[width=0.31\textwidth]{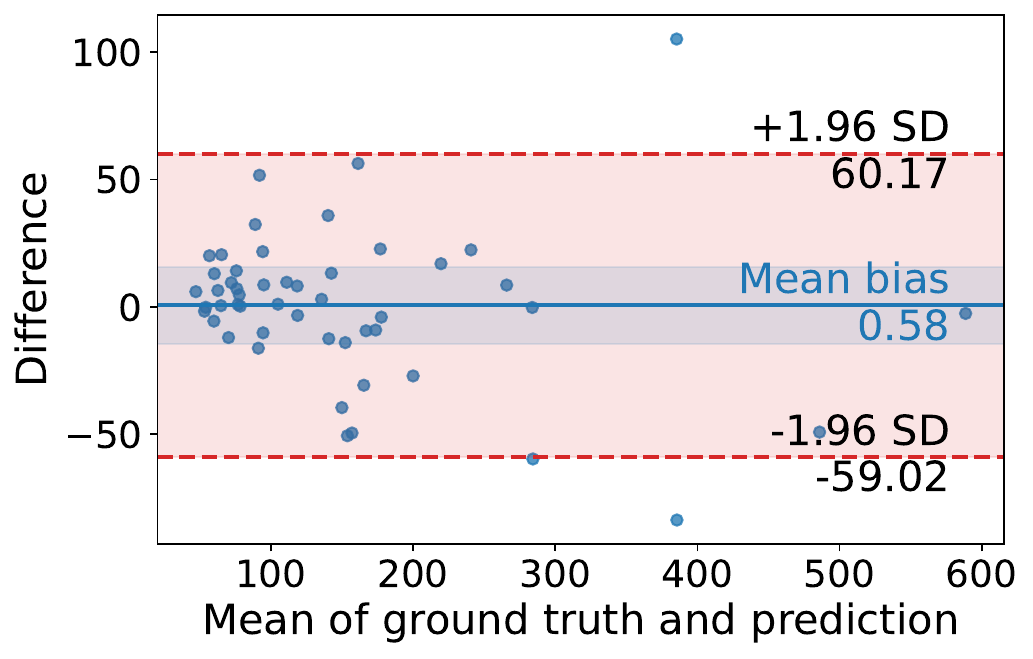}}\hfill
\subfloat[EF of our PROMISE-T\textsubscript{xformer}]{\includegraphics[width=0.31\textwidth]{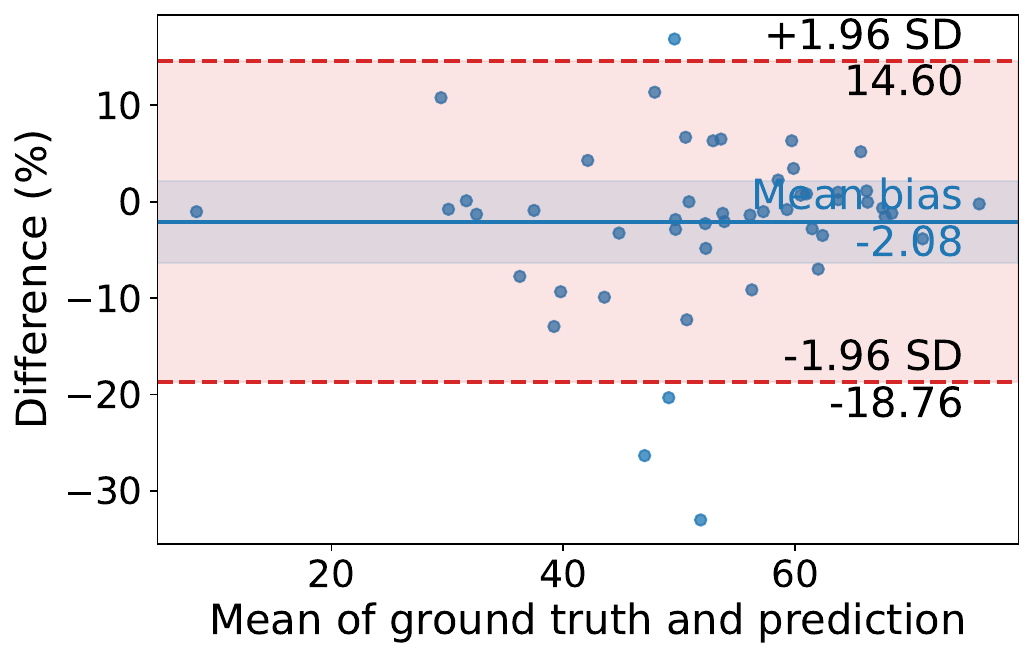}}

\caption{Bland--Altman plots comparing predicted and reference EDV, ESV, and EF for UNETR (a-c) and our PROMISE-T\textsubscript{xformer} (d-f). The proposed model reduced the mean bias (EDV: $-13.9 \rightarrow -6.7$~mL; EF: $-4.15 \rightarrow -2.08$\%) and narrowed the LoA across all indices, demonstrating improved volumetric precision and enhanced temporal coherence. The EF agreement ($-18.8$\%~to~$+14.6$\%) lies within the inter-observer variability range, confirming clinically reliable ventricular function estimation.
}
\label{fig:EFs}
\end{figure*}

\subsection{Quantitative Agreement in Ventricular Volumes and Function}
To assess quantitative agreement between estimated and reference measurements, Bland-Altman analyses were conducted for EDV, ESV, and EF on the CAMUS dataset (see Fig.~\ref{fig:EFs}). 
The baseline UNETR exhibited a systematic underestimation of EDV (mean bias $=-13.9$~mL, $95\%$ LoA $=[-131.4,\,+103.5]$~mL) and wide LoA, indicating large inter-subject dispersion, particularly for subjects with dilated ventricles. 
The proposed PROMISE-T\textsubscript{xformer} reduced this bias to $-6.7$~mL and narrowed the LoA to $[-95.3,\,+81.9]$~mL, reflecting more accurate delineation of the endocardial surface at ED and improved volumetric consistency. 
For ESV, both models produced negligible bias ($\approx0.6$~mL), but PROMISE-T\textsubscript{xformer} achieved tighter LoA ($\pm60$~mL vs.\ $\pm71$~mL), suggesting higher precision during the contracted phase. 
The improvement was most evident for EF (improving the correlation from 72.9\% to 80.2\%), where the mean bias decreased from $-4.15\%$ to $-2.08\%$, and the LoA contracted from $[-23.9\%,\,+15.6\%]$ to $[-18.8\%,\,+14.6\%]$. 
The smaller EF dispersion indicates enhanced temporal coherence between ED and ES segmentations, attributable to the phase-consistent contextual attention mechanism. 
No proportional bias was observed ($|r|<0.1$, $p>0.05$), confirming stable performance across the physiological range of ventricular sizes and functions. 
The EF limits fall within the reported inter-observer variability of expert annotations ($\sim\pm15\%$), indicating that PROMISE-T\textsubscript{xformer} provides clinically interchangeable functional estimation. 
Collectively, these findings demonstrate that integrating interactive prompts through PCCA attention enhances not only segmentation fidelity but also the reproducibility of physiologically meaningful volumetric and functional indices.

\begin{figure*}[!h]
\centering
\includegraphics[width=\linewidth]{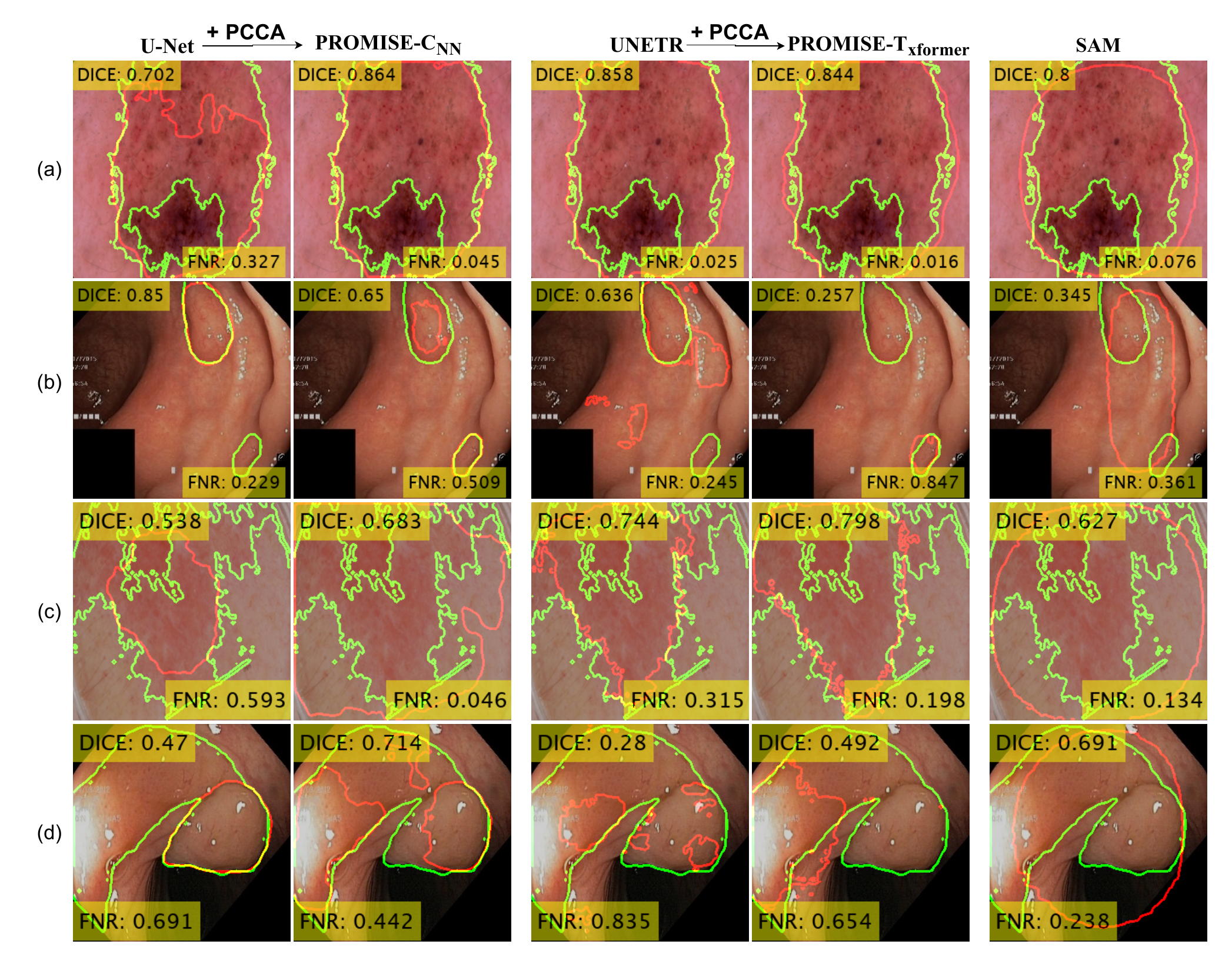}
\caption{Qualitative analysis of challenging and failure cases on ISIC-2017 and Kvasir-Polyp. Each column compares U-Net, PROMISE-C\textsubscript{NN}, UNETR, PROMISE-T\textsubscript{xformer}, and SAM. Low-contrast textures, disconnected regions, and complex boundary structures lead to degraded performance across all methods; however, PROMISENet variants consistently preserve better boundary conformity than baseline U-Net and UNETR under extreme cases.}
\label{fig:failure_modes}
\end{figure*}

\subsection{Benchmark with State-of-the-Art}
We benchmark PROMISE-Net against a diverse set of recent state-of-the-art methods, encompassing both convolutional and transformer-based architectures, including Pact-Net~\cite{chen2023pact}, USL-Net~\cite{li2024usl}, FAT-Net~\cite{wu2022fat}, HTC-Net~\cite{tang2024htc}, RM-U-Net~\cite{tang2024rm}, ERDUNet~\cite{li2023erdunet}, MAF-Net~\cite{yang2023maf}, DECA-Net~\cite{liang2024deca}, CLAS~\cite{10.1007/978-3-030-59713-9_60}, TAM-FCN8s~\cite{hasan2025motion}, and CoST-UNet~\cite{ISLAM2024106633}. All methods are evaluated on the same datasets, as summarized in Table~\ref{tab:TAM_SAM4}.

Across all datasets, PROMISE-C\textsubscript{NN} achieves the highest or near-highest segmentation accuracy while simultaneously attaining the lowest boundary deviation (HD95) and false-negative rate (FNR). On ISIC-2017, it reaches a DSC of 90.7\% and an FNR of 7.3\%, outperforming HTC-Net by 0.6\% in DSC and reducing FNR by approximately 38\% (from 11.8\% to 7.3\%). On CAMUS, PROMISE-C\textsubscript{NN} attains a DSC of 91.3\% and an IoU of 84.4\%, outperforming CC-SAM and TAM-FCN8s, while reducing HD95 to 5.9 pixels, indicating improved boundary precision in low-contrast echocardiographic images. Similar gains are observed on Kvasir-Polyp and Kvasir-Instrument, where PROMISE-C\textsubscript{NN} consistently outperforms alternative methods (Table~\ref{tab:TAM_SAM4}). Overall, it yields an average DSC improvement of approximately 2--3\% and reduces FNR by up to 35\% relative to the strongest competitors, demonstrating that prompt-conditioned channel modulation reliably enhances segmentation accuracy and boundary fidelity across diverse anatomies and imaging modalities.

\subsection{Failure Analysis and Limitations}
We further analyse failure cases on ISIC-2017 and Kvasir-Polyp (Fig.~\ref{fig:failure_modes}). While PROMISE-Net variants consistently improve contour smoothness and reduce false negatives, residual limitations remain under visually ambiguous or structurally irregular conditions.

In ISIC-2017 (Fig.~\ref{fig:failure_modes}(a)), lesions with internal hollows or heterogeneous pigmentation produce ambiguous intensity gradients that hinder precise boundary localization. Even with prompt-conditioned modulation, internal textures may be conflated with lesion boundaries, leading to mild over-segmentation. This suggests that future designs could benefit from spatial-frequency–aware attention or texture-decoupled feature modulation to better preserve fine internal contrast. Figure~\ref{fig:failure_modes}(b) shows a Kvasir-Polyp failure case in which a single image contains multiple spatially disconnected polyps. Here, a single global prompt or bounding box is insufficient to attend to multiple targets simultaneously, resulting in missed or partially segmented regions. More challenging cases in Fig.~\ref{fig:failure_modes}(c)–(d) exhibit highly irregular boundaries, fragmented structures, and specular highlights that introduce strong local ambiguities. Although PROMISE-Net improves boundary adherence relative to baseline and foundation models in these cases, it still struggles to fully capture such complex contours. This motivates future extensions of PROMISE-Net with multi-scale prompt reasoning, adaptive multi-prompting, uncertainty modeling, and richer spatial context aggregation.

\section{Conclusion and Future Extensions}
This paper introduced PROMISE-Net, a prompt-aware medical image segmentation framework built on the proposed PCCA mechanism. By embedding prompt-conditioned modulation hierarchically across encoder and decoder stages, PROMISE-Net enables spatial prompts to influence feature representations at multiple semantic levels rather than being restricted to late-stage fusion. Extensive experiments across four heterogeneous benchmarks (ISIC-2017, Kvasir-SEG, Kvasir-Instrument, and CAMUS) demonstrate that PROMISE-Net delivers consistent improvements in overlap accuracy, boundary precision, and false-negative reduction over strong convolutional, transformer-based, and prompt-driven baselines. The results further confirm robust generalization across architectures (PROMISE-C\textsubscript{NN} and PROMISE-T\textsubscript{xformer}), anatomical targets, imaging modalities, image quality levels, and label-space complexity (binary vs.\ multi-class segmentation), while remaining resilient to inter-observer variability and prompt perturbations.

Beyond segmentation accuracy, PROMISE-Net shows improved anatomical consistency and clinically meaningful reliability, as evidenced by reduced boundary errors, stable performance under degraded image quality, and improved agreement in downstream cardiac volumetric and functional measurements. Collectively, these findings establish prompt-conditioned channel modulation as an effective and generalizable strategy for interactive medical image segmentation.

While this work focuses on 2D medical image segmentation, the modular design of PCCA readily supports several important extensions. First, the hierarchical prompt-conditioning mechanism can be naturally extended to 3D and 4D (spatio-temporal) segmentation, enabling motion-aware analysis of volumetric and dynamic data such as cardiac cycles, fetal echocardiography, or endoscopic video streams. Second, the prompt encoder can be generalized to support multi-prompt interactions, including combinations of bounding boxes, points, and scribbles, allowing finer control over ambiguous or overlapping anatomical regions with minimal user effort. 

In addition, integrating domain adaptation and continual learning strategies would further strengthen cross-institutional robustness, facilitating deployment across scanners, centers, and patient populations. Finally, extending prompt-conditioned modulation to multimodal settings, such as joint ultrasound–MRI or image–clinical metadata fusion, represents a promising direction for building context-aware and clinically scalable segmentation systems. 

Overall, PROMISE-Net provides a flexible foundation for advancing interactive, prompt-driven, and anatomically reliable segmentation, with clear pathways toward higher-dimensional, multimodal, and real-time clinical applications.

\section*{Declaration of Competing Interest}
The authors declare that they have no known competing financial interests or personal relationships that could have appeared to influence the work reported in this paper.

\bibliographystyle{model1-num-names}
\bibliography{sample}

\end{document}